\documentclass{article}

\usepackage[preprint,nonatbib]{neurips_2019}

\usepackage[utf8]{inputenc} 
\usepackage[T1]{fontenc}    
\usepackage{url}            
\usepackage{booktabs}       
\usepackage{amsfonts}       
\usepackage{nicefrac}       
\usepackage{microtype}      

\usepackage[round]{natbib}
\usepackage{xcolor}
\usepackage{times}
\usepackage{soul}
\usepackage[small]{caption}
\usepackage{graphicx}
\usepackage{amsmath}
\usepackage{xargs}
\usepackage{amsthm}
\usepackage{mathtools}
\usepackage{dsfont}
\usepackage{import}
\usepackage{svg}
\usepackage{pdfpages}
\usepackage[norelsize, linesnumbered, ruled, lined, boxed, commentsnumbered]{algorithm2e}
\usepackage{hyperref}

\definecolor{mylinkcolor}{HTML}{e83f6f}
\definecolor{mycitecolor}{HTML}{2980B9}
\definecolor{myurlcolor}{HTML}{d63230}
\definecolor{myalgocolor}{HTML}{000000}

\hypersetup{
  linkcolor  = mylinkcolor,
  citecolor  = mycitecolor,
  urlcolor   = myurlcolor,
  colorlinks = true,
}

\usepackage{cleveref}
\usepackage{float}
\usepackage{tabularx}
\usepackage{makecell}
\usepackage{wrapfig}
\usepackage{enumerate}
\usepackage{listings}
\newcommand{\FTQ}{\textcolor{myalgocolor}{\normalfont \texttt{FTQ}}\xspace}
\newcommand{\FTQl}{\textcolor{myalgocolor}{\normalfont \texttt{FTQ}$(\lambda)$}\xspace}
\newcommand{\BFTQ}{\textcolor{myalgocolor}{\normalfont \texttt{BFTQ}}\xspace}
\newcommand{\ov}{\overline}
\newcommand{\oa}{\ov{a}}

\newcommand{\os}{\ov{s}}
\newcommand{\ocS}{\ov{\cS}}
\newcommand{\ocA}{\ov{\cA}}

\renewcommand{\epsilon}{\varepsilon}
\renewcommand{\tilde}{\widetilde}

\newcommand*{\MyDef}{\mathrm{\tiny def}}
\newcommand*{\eqdefU}{\ensuremath{\mathop{\overset{\MyDef}{=}}}}
\newcommand*{\eqdef}{\mathop{\overset{\MyDef}{\resizebox{\widthof{\eqdefU}}{\heightof{=}}{=}}}}

\def\:#1{\protect \ifmmode {\mathbf{#1}} \else {\textbf{#1}} \fi}

\newcommand{\cA}{\mathcal{A}}
\newcommand{\cB}{\mathcal{B}}
\newcommand{\cC}{\mathcal{C}}
\newcommand{\cD}{\mathcal{D}}

\newcommand{\cF}{\mathcal{F}}

\newcommand{\cH}{\mathcal{H}}

\newcommand{\cL}{\mathcal{L}}
\newcommand{\cM}{\mathcal{M}}

\newcommand{\cO}{\mathcal{O}}
\newcommand{\cP}{\mathcal{P}}

\newcommand{\cS}{\mathcal{S}}
\newcommand{\cT}{\mathcal{T}}
\newcommand{\cU}{\mathcal{U}}

\renewcommand{\epsilon}{\varepsilon}

\DeclareMathOperator*{\argmin}{arg\,min} 
\DeclareMathOperator*{\argmax}{arg\,max}

\newcommand{\probability}[1]{\mathbb{P}\left(#1\right)}

\DeclareMathOperator*{\expectedvalue}{\mathbb{E}}

\newcommand{\expectedvalueover}[1]{\expectedvalue\limits_{#1}}
\newcommand{\condbar}{\;\middle|\;}

\newcommand{\Real}{\mathbb{R}}
\newcommand{\Natural}{\mathbb{N}}

\newtheorem{theorem}{Theorem}
\newtheorem{definition}{Definition}

\newtheorem{proposition}{Proposition}
\newtheorem{remark}{Remark}

\usepackage[colorinlistoftodos,prependcaption,textsize=tiny]{todonotes}
\usetikzlibrary{patterns}

\newcommandx{\nc}[2][1=]{\todo[linecolor=blue,backgroundcolor=blue!25,bordercolor=blue,#1]{\textbf{Nicolas:} #2}}
\newcommandx{\op}[2][1=]{\todo[linecolor=green,backgroundcolor=green!25,bordercolor=green,#1]{\textbf{Olivier:} #2}}
\newcommandx{\el}[2][1=]{\todo[linecolor=red,backgroundcolor=red!25,bordercolor=red,#1]{\textbf{Edouard:} #2}}
\newcommandx{\td}[2][1=]{\todo[inline,size=\large,#1]{#2}}

\makeatletter
\providecommand*{\input@path}{}
\g@addto@macro\input@path{{./source/}{./source/ressources/}{./ressources/}}
\makeatother

\crefname{algocf}{alg.}{algs.}
\Crefname{algocf}{Algorithm}{Algorithms}

\title{Budgeted Reinforcement Learning in Continuous State Space}

\author{
Nicolas Carrara\thanks{Both authors contributed equally.}\\
SequeL team, INRIA Lille -- Nord Europe\thanks{Univ. Lille, CNRS, Centrale Lille, INRIA UMR 9189 - CRIStAL, Lille, France}\\
\texttt{nicolas.carrara@inria.fr}
\And
Edouard Leurent\footnotemark[1]\\
SequeL team, INRIA Lille -- Nord Europe\footnotemark[2]\\
Renault Group, France\\
\texttt{edouard.leurent@inria.fr}
\And
Romain Laroche\\
Microsoft Research, Montreal, Canada\\
\texttt{romain.laroche@microsoft.com}
\And
Tanguy Urvoy\\
Orange Labs, Lannion, France\\
\texttt{tanguy.urvoy@orange.com}
\And
Odalric-Ambrym Maillard\\
SequeL team, INRIA Lille -- Nord Europe\\
\texttt{odalric.maillard@inria.fr}
\And
Olivier Pietquin\\
Google Research - Brain Team\\
SequeL team, INRIA Lille -- Nord Europe\footnotemark[2]\\
\texttt{pietquin@google.com}
}

\begin{document}
\maketitle
\begin{abstract}
    A Budgeted Markov Decision Process (BMDP) is an extension of a Markov Decision Process to critical applications requiring safety constraints. It relies on a notion of risk implemented in the shape of a cost signal constrained  to lie below an -- adjustable -- threshold.
    So far, BMDPs could only be solved in the case of finite state spaces with known dynamics. This work extends the state-of-the-art to continuous spaces environments and unknown dynamics. We show that the solution to a BMDP is a fixed point of a novel Budgeted Bellman Optimality operator. This observation allows us to introduce natural extensions of Deep Reinforcement Learning algorithms to address large-scale BMDPs. We validate our approach on two simulated applications: spoken dialogue and autonomous driving.
\end{abstract}

\section{Introduction}
\label{sec:intro}

Reinforcement Learning (RL) is a general framework for decision-making under uncertainty. It frames the learning objective as the optimal control of a Markov Decision Process  $(\cS, \cA, P, R_r, \gamma)$ with measurable state space $\cS$, discrete actions $\cA$, unknown rewards $R_r\in\Real^{\cS \times \cA}$, and unknown dynamics $P\in \cM(\cS)^{\cS \times \cA}$ , where $\cM(\mathcal{X})$ denotes the probability measures over a set $\mathcal{X}$. Formally, we seek a policy $\pi\in\cM(A)^\cS$ that maximises in expectation the $\gamma$-discounted return of rewards $G_r^\pi = \sum_{t=0}^\infty \gamma^t R_r(s_t, a_t)$.

However, this modelling assumption comes at a price: no control is given over the spread of the performance distribution \citep{Dann2018}. In many critical real-world applications where failures may turn out very costly, this is an issue as most decision-makers would rather give away some amount of expected optimality to increase the performances in the lower-tail of the distribution. This has led to the development of several risk-averse variants where the optimisation criteria include other statistics of the performance, such as the worst-case realisation \citep{Iyengar2005,Nilim2005,Wiesemann2013}, the variance-penalised expectation \citep{Garcia2015,Tamar2012}, the Value-At-Risk (VaR) \citep{Mausser2003,Luenberger2013}, or the Conditional Value-At-Risk (CVaR) \citep{Chow2014,ChowGJP15}.

Reinforcement Learning also assumes that the performance can be described by a single reward function $R_r$. Conversely, real problems typically involve many aspects, some of which can be contradictory \citep{Liu2014}. For instance, a self-driving car needs to balance between progressing quickly on the road and avoiding collisions. When aggregating several objectives in a single scalar signal, as often in Multi-Objectives RL~\citep{Roijers2013ASO}, no control is given over their relative ratios, as high rewards can compensate high penalties. For instance, if a weighted sum is used to balance velocity $v$ and crashes $c$, then for any given choice of weights $\omega$ the optimality equation $\omega_v\expectedvalue[\sum\gamma^t v_t] + \omega_a\expectedvalue[\sum\gamma^t c_t] = G_r^* = \max_\pi G^\pi_r$ is the equation of a line in $(\expectedvalue[\sum\gamma^t v_t], \expectedvalue[\sum\gamma^t c_t])$, and the automotive company cannot control where its optimal policy $\pi^*$ lies on that line.

Both of these concerns can be addressed in the \emph{Constrained Markov Decision Process} (CMDP) setting \citep{BEUTLER1985236,Altman95constrainedmarkov}. In this multi-objective formulation, task completion and safety are considered separately. We equip the MDP with a cost signal $R_c \in \Real^{\cS\times \cA}$ and a cost budget $\beta\in\Real$. Similarly to $G_r^\pi$, we define the return of costs $G_c^\pi = \sum_{t=0}^\infty \gamma^t R_c(s_t, a_t)$ and the new cost-constrained objective:
\begin{equation}
\label{eq:cmdp}
\begin{array}{lcr}
 \displaystyle \max_{\pi\in\cM(\cA)^\cS} \expectedvalue[G_r^\pi | s_0=s] & \text{ s.t. } & \expectedvalue[G_c^\pi | s_0=s] \leq \beta
\end{array}
\end{equation}

This constrained framework allows for better control of the performance-safety tradeoff. However, it suffers from a major limitation: the budget has to be chosen before training, and cannot be changed afterwards.

To address this concern, the \emph{Budgeted Markov Decision Process} (BMDP) was introduced in \citep{Boutilier_Lu:uai16} as an extension of CMDPs to enable the online control over the budget $\beta$ within an interval $\cB \subset \Real$ of admissible budgets. Instead of fixing the budget prior to training, the objective is now to find a generic optimal policy $\pi^*$ that takes $\beta$ as input so as to solve the corresponding CMDP (Eq. \eqref{eq:cmdp}) for all $\beta\in\cB$. This gives the system designer the ability to move the optimal policy $\pi^*$ in real-time along the Pareto-optimal curve of the different reward-cost trade-offs.

Our first contribution is to re-frame the original BMDP formulation in the context of continuous states and infinite discounted horizon. We then propose a novel Budgeted Bellman Optimality Operator and prove the optimal value function to be a fixed point of this operator. Second, we use this operator in \BFTQ, a batch Reinforcement Learning algorithm, for solving BMDPs online by interaction with an environment, through function approximation and a tailored exploration procedure. Third, we scale this algorithm to large problems by providing an efficient implementation of the Budgeted Bellman Optimality Operator based on convex programming, and by leveraging tools from Deep Reinforcement Learning such as Deep Neural Networks and synchronous parallel computing. Finally, we validate our approach in two environments that display a clear trade-off between rewards and costs: a spoken dialogue system and a problem of behaviour planning for autonomous driving. The proofs of our main results are provided in \Cref{sec:proofs}.

\section{Budgeted Dynamic Programming}
\label{sec:bdp}
 We work in the space of budgeted policies, where $\pi$ both depends on $\beta$ and also outputs a next budget $\beta_a$. Hence, the budget $\beta$ is neither fixed nor constant as in the CMDP setting but instead evolves as part of the dynamics.

We cast the BMDP problem as a multi-objective MDP problem \citep{Roijers2013ASO} by considering \emph{augmented} state and action spaces $\ocS = \cS\times \cB$ and $\ocA= \cA\times \cB$, and equip them with the augmented dynamics $\ov{P}\in \cM(\ocS)^{\ocS \times \ocA}$ defined as:
\begin{equation}
    \label{eq:dynamics}
    \ov{P}\left(\os' \condbar \os, \oa\right) = \ov{P}\left((s',\beta') \condbar (s,\beta), (a, \beta_a)\right) \eqdef P(s'|s, a)\delta(\beta' - \beta_a),
\end{equation}
where $\delta$ is the Dirac indicator distribution.

In other words, in these augmented dynamics, the output budget $\beta_a$ returned at time $t$ by a budgeted policy $\pi\in \Pi=\cM(\ocA)^{\ocS}$ will be used to condition the policy at the next timestep $t+1$.

We stack the rewards and cost functions in a single \emph{vectorial} signal $R \in (\Real^2)^{{\ocS \times \ocA}}$.
Given an augmented transition $(\os, \oa) =((s,\beta), (a, \beta_a))$, we define:
\begin{equation}
     R(\os, \oa) \eqdef  \begin{bmatrix}
     R_r(s, a)\\
     R_c(s, a)
     \end{bmatrix}\in\Real^2.
\end{equation}

Likewise, the return $G^\pi = (G_r^\pi, G_c^\pi)$ of a budgeted policy $\pi\in\Pi$ refers to:
$G^\pi \eqdef \sum_{t=0}^\infty \gamma^t R(\os_t, \oa_t)$,
and the value functions $V^\pi$, $Q^\pi$ of a budgeted policy $\pi\in\Pi$ are defined as:
\begin{equation}
    \label{eq:value-function}
V^\pi(\os) = (V_r^\pi, V_c^\pi) \eqdef \expectedvalue\left[ G^\pi \condbar \ov{s_0} = \os\right] \qquad Q^\pi(\os, \oa)= (Q_r^\pi, Q_c^\pi) \eqdef \expectedvalue\left[ G^\pi \condbar \ov{s_0} = \os, \ov{a_0} = \oa\right].
\end{equation}
We restrict $\ocS$ to feasible budgets only: $\ocS_f \eqdef \{(s,\beta)\in\ocS: \exists \pi\in\Pi, V_c^\pi(s) \geq \beta\}$ that we assume is non-empty for the BMDP to admit a solution. We still write $\ocS$ in place of $\ocS_f$ for brevity of notations.

\begin{proposition}[Budgeted Bellman Expectation]
\label{prop:bellman-expectation}
The value functions $V^\pi$ and $ Q^\pi$ verify:
\begin{align}
    V^\pi(\os) = \sum_{\oa\in\ocA}\pi(\oa | \os) Q^\pi(\os, \oa) \qquad Q^\pi(\os, \oa) = R(\os, \oa) + \gamma\sum_{\os'\in\ocS}\ov{P}\left(\os' \condbar \os, \oa\right) V^\pi(\os') \label{eq:bellman_expectation}
\end{align}
Moreover, consider the Budgeted Bellman Expectation operator $\cT^\pi$:
$\forall Q\in(\Real^2)^{\ocS\ocA}, \os\in\ocS, \oa\in\ocA$,
\begin{align}
\label{eq:bellman_expectation_operator}
    \cT^\pi Q(\os, \oa) &\eqdef R(\os, \oa) + \gamma \sum_{\os'\in\ocS}\sum_{\oa'\in\ocA}\ov{P}(\os'|\os, \oa)\pi(\oa'|\os') Q(\os', \oa')
\end{align}
Then $\cT^\pi$ is a $\gamma$-contraction and $Q^\pi$ is its unique fixed point.
\end{proposition}

\begin{definition}[Budgeted Optimality]
We now come to the definition of budgeted optimality. We want an optimal budgeted policy to: (i)~respect the cost budget $\beta$, (ii)~maximise the $\gamma$-discounted return of rewards $G_r$, (iii)~in case of tie, minimise the $\gamma$-discounted return of costs $G_c$. To that end, we define for all $\os\in\ocS$:
\begin{enumerate}[(i)]
    \item Admissible policies $\Pi_a$: 
    \begin{equation}
    \Pi_a(\os) \eqdef \{\pi\in\Pi: V_c^\pi(\os) \leq \beta\}\text{ where }\os = (s, \beta)
    \end{equation}
    \item Optimal value function for rewards $V_r^*$ and candidate policies $\Pi_r$: 
    \begin{equation}
        V_r^*(\os) \eqdef \max_{\pi\in\Pi_a(\os)}  V_r^\pi(\os) \qquad\qquad \Pi_r(\os) \eqdef \argmax_{\pi\in\Pi_a(\os)}  V_r^\pi(\os)
    \end{equation}
    \item Optimal value function for costs $V_c^*$ and optimal policies $\Pi^*$: 
    \begin{equation}
        V_c^*(\os) \eqdef \min_{\pi\in\Pi_r(\os)}  V_c^\pi(\os), \qquad\qquad \Pi^*(\os) \eqdef \argmin_{\pi\in\Pi_r(\os)}  V_c^\pi(\os)
    \end{equation}
\end{enumerate}
We define the budgeted action-value function $Q^*$ similarly:
\begin{equation}
    Q_r^*(\os, \oa) \eqdef \max_{\pi\in\Pi_a(\os)}  Q_r^\pi(\os, \oa) \qquad\qquad Q_c^*(\os, \oa) \eqdef \min_{\pi\in\Pi_r(\os)}  Q_c^\pi(\os, \oa) 
\end{equation}
and denote $V^* = (V_r^*, V_c^*)$, $Q^* = (Q_r^*, Q_c^*)$.
\end{definition}

\begin{theorem}[Budgeted Bellman Optimality]
\label{thm:bellman-optimality}
The optimal budgeted action-value function $Q^*$ verifies:
\begin{equation}
\label{eq:bellman-optimality}
    Q^{*}(\os, \oa) = \cT Q^{*}(\os, \oa) \eqdef R(\os, \oa) + \gamma \sum_{\os'\in\ocS} \ov{P}(\ov{s'} | \os, \oa)\sum_{\ov{a'}\in \ocA} \pi_\text{greedy}(\ov{a'}|\ov{s'}; Q^*) Q^{*}(\ov{s'}, \ov{a'}),
\end{equation}
where the greedy policy $\pi_\text{greedy}$ is defined by: $\forall \os=(s,\beta)\in \ocS, \oa\in 
\ocA, \forall Q\in(\Real^2)^{\ocA\times\ocS},$
\begin{subequations}
\label{eq:pi_greedy}
\begin{align}
    \pi_\text{greedy}(\oa|\os; Q) \in &\argmin_{\rho\in\Pi_r^Q} \expectedvalueover{\oa\sim\rho}Q_c(\os, \oa), \label{eq:pi_greedy_cost}\\
    \text{where }\quad\Pi_r^Q \eqdef &\argmax_{\rho\in\cM(\ocA)} \expectedvalueover{\oa\sim\rho} Q_r(\os, \oa) \label{eq:pi_greedy_reward}\\
    & \text{ s.t. }  \expectedvalueover{\oa\sim\rho} Q_c(\os, \oa) \leq \beta. \label{eq:pi_greedy_constraint}
\end{align}
\end{subequations}
\end{theorem}

\begin{remark}[Appearance of the greedy policy]
\label{rmk:greedy}
In classical Reinforcement Learning, the greedy policy takes a simple form $\pi_\text{greedy}(s; Q^*) = \argmax_{a\in\cA} Q^*(s, a)$, and the term $\pi_\text{greedy}(a'|s';Q^*) Q^{*}(s', a')$ in \eqref{eq:bellman-optimality} conveniently simplifies to $\max_{a'\in \cA} Q^*(s', a')$. Unfortunately, in a budgeted setting the greedy policy requires solving the nested constrained optimisation program \eqref{eq:pi_greedy} at each state and budget in order to apply this Budgeted Bellman Optimality operator.
\end{remark}

\begin{proposition}[Optimality of the greedy policy]
\label{prop:greedy_optimal}
The greedy policy $\pi_\text{greedy}(\cdot~; Q^*)$ is \emph{uniformly} optimal: for all $\os\in\ocS$, $\pi_\text{greedy}(\cdot~; Q^*)\in\Pi^*(\os)$. In particular, $V^{\pi_\text{greedy}(\cdot; Q^*)} = V^*$ and $Q^{\pi_\text{greedy}(\cdot; Q^*)}= Q^*$.
\end{proposition}

\paragraph{Budgeted Value Iteration}

The Budgeted Bellman Optimality equation is a fixed-point equation, which motivates the introduction of a fixed-point iteration procedure. We introduce \Cref{algo:bvi}, a Dynamic Programming algorithm for solving known BMDPs. If it were to converge to a unique fixed point, this algorithm would provide a way to compute $Q^*$ and recover the associated optimal budgeted policy $\pi_\text{greedy}(\cdot~; Q^*)$.

\begin{theorem}[Non-contractivity of $\cT$]
\label{thm:contraction}
For any BMDP ($\cS,\cA,P,R_r,R_c,\gamma$) with $|\cA| \geq 2$, $\cT$ is not a contraction. Precisely: $\forall\epsilon>0, \exists Q^1,Q^2\in(\Real^2)^{\ocS\ocA}:\|\cT Q^1-\cT Q^2\|_\infty \geq \frac{1}{\epsilon}\|Q^1-Q^2\|_\infty$.
\end{theorem}

Unfortunately, as $\cT$ is not a contraction, we can guarantee neither the convergence of \Cref{algo:bvi} nor the unicity of its fixed points. Despite those theoretical limitations, we empirically observed the convergence to a fixed point in our experiments (\Cref{sec:experiements}). We conjecture a possible explanation:

\begin{remark}[Contractivity of $\cT$ on smooth $Q$-functions]
\label{rmk:contractivity-smooth}
We conjecture that $\cT$ is a contraction when restricted to the subset $\cL_\gamma$ of $Q$-functions such that "$Q_r$ is $L$-Lipschitz with respect to $Q_c$", with $L<\frac{1}{\gamma}-1$. We lengthily discuss some intuition on why that should be the case in \Cref{proof:contraction-with-smooth}.
\end{remark}


\section{Budgeted Reinforcement Learning}
\label{sec:brl}

In this section, we consider BMDPs with unknown parameters that must be solved by interaction with an environment. 

\subsection{Budgeted Fitted-Q}
\label{subsec:bftq}

When the BMDP is unknown, we need to adapt \Cref{algo:bvi} to work with a batch of samples $\cD=\{(\os_i,\oa_i,r_i,\os_i'\}_{i\in [0,N]}$ collected by interaction with the environment. Applying $\cT$ in \eqref{eq:bellman-optimality} would require computing an expectation $\expectedvalue_{\os'\sim \ov{P}}$ over next states $\os'$ and hence an access to the model $\ov{P}$. We instead use $\hat{\cT}$, a sampling operator, in which this expectation is replaced by:
\begin{equation*}
    \hat{\cT} Q(\os_i, \oa_i, r_i, \os'_i) \eqdef r_i + \gamma \sum_{\ov{a'_i}\in \cA_i} \pi_\text{greedy}(\ov{a'_i}|\ov{s'_i}; Q) Q(\ov{s'_i}, \ov{a'_i}).
\end{equation*}
We introduce in \Cref{algo:bftq} the \emph{Budgeted-Fitted-Q} (\BFTQ) algorithm, an extension of the \emph{Fitted-Q} (\FTQ) algorithm \citep{Ernst2005,Riedmiller2005} adapted to solve unknown BMDPs. Because we work with  continuous state space $\cS$ and budget space $\cB$, we need to employ function-approximation in order to generalise to nearby states and budgets. Precisely, given a parametrized model $Q_\theta$, we seek to minimise a regression loss $\cL(Q_\theta, Q_\text{target};\cD) = \sum_{\cD} ||Q_\theta(\os, \oa) - Q_\text{target}(\os, \oa, r, \os')||_2^2$.
Any model can be used, such as linear models, regression trees, or neural networks.

\begin{minipage}[t]{0.47\textwidth}
\vspace{0pt}  

\begin{algorithm}[H]
\DontPrintSemicolon
\KwData{$P, R_r, R_c$}
\KwResult{$Q^*$}
$Q_{0} \leftarrow 0$\;
\Repeat{convergence}{
    $Q_{k+1} \leftarrow \cT Q_k$\;
}
\caption{Budgeted Value Iteration}
\label{algo:bvi}

\end{algorithm}

\end{minipage}%
\hfill
\begin{minipage}[t]{0.47\textwidth}
\vspace{0pt}

\begin{algorithm}[H]
\DontPrintSemicolon
\KwData{$\cD$}
\KwResult{$Q^*$}
$Q_{\theta_0} \leftarrow 0$\;
\Repeat{convergence}{
    $\theta_{k+1} \leftarrow \argmin_\theta \cL(Q_\theta, \hat{\cT} Q_{\theta_{k}}; \cD)$\;
}
\caption{Budgeted Fitted-Q}
\label{algo:bftq}

\end{algorithm}

\end{minipage}

\subsection{Risk-sensitive exploration}
\label{sec:exploration}

In order to run \Cref{algo:bftq}, we must first gather a batch of samples $\mathcal{D}$. Ideally we would need samples from the asymptotic state-budget distribution $\lim_{t\rightarrow\infty}\probability{\os_t}$ induced by an optimal policy $\pi^*$ given an initial distribution $\probability{\os_0}$, but as we are actually building this policy, it is not possible. Following the same idea of $\epsilon$-greedy exploration for \FTQ \citep{Ernst2005,Riedmiller2005}, we introduce an algorithm for risk-sensitive exploration. We follow an exploration policy: a mixture between a random budgeted policy $\pi_\text{rand}$ and the current greedy policy $\pi_\text{greedy}$. The batch $\cD$ is split into several mini-batches generated sequentially, and $\pi_\text{greedy}$ is updated by running \Cref{algo:bftq} on $\cD$ upon mini-batch completion. $\pi_\text{rand}$ is designed to obtain trajectories that only explore feasible budgets: we impose that the joint distribution $\probability{a, \beta_a|s, \beta}$ verifies $\expectedvalue[\beta_a]\leq\beta$. This condition defines a probability simplex $\Delta_{\ocA}$ from which we sample uniformly. Finally, when interacting with an environment the initial state $s_0$ is usually sampled from a starting distribution $\probability{s_0}$. In the budgeted setting, we also need to sample the initial budget $\beta_0$. Importantly, we pick a uniform distribution $\probability{\beta_0} = \cU(\cB)$ so that the entire range of risk-level is explored, and not only reward-seeking behaviours as would be the case with a traditional risk-neutral $\epsilon$-greedy strategy. The pseudo-code of our exploration procedure is shown in \Cref{algo:risk-sensitive-exploration} in \Cref{sec:risk-sensitive-supp}.

\section{A Scalable Implementation}
\label{sec:scalable-bftq}
In this section, we introduce an implementation of the \BFTQ algorithm designed to operate efficiently and handle large batches of experiences $\mathcal{D}$.

\subsection{How to compute the greedy policy?}
\label{subsec:compute-greedy-policy}
As stated in \Cref{rmk:greedy}, computing the greedy policy $\pi_\text{greedy}$ in \eqref{eq:bellman-optimality} is not trivial since it requires solving the nested constrained optimisation program \eqref{eq:pi_greedy}.
However, it can be solved efficiently by exploiting the \emph{structure} of the set of solutions with respect to $\beta$, that is, concave and increasing. 

\begin{proposition}[Equality of $\pi_\text{greedy}$ and $\pi_\text{hull}$]
\label{prop:bftq_pi_hull}
\Cref{algo:bvi} and \Cref{algo:bftq} can be run by replacing $\pi_\text{greedy}$ in the equation \eqref{eq:bellman-optimality} of $\cT$ with $\pi_\text{hull}$ as described in \Cref{algo:pi_hull}.
\begin{equation*}
    \pi_\text{greedy}(\oa|\os; Q) = \pi_\text{hull}(\oa|\os; Q)
\end{equation*}
\end{proposition}

\begin{algorithm}
\DontPrintSemicolon
\KwData{$\os=(s,\beta)$, $Q$}
$Q^+\leftarrow \{Q_c > \min \{Q_c(\os,\oa) \text{ s.t. }\oa\in\argmax_{\oa} Q_r(\os,\oa)\} \}$\tcp*[f]{dominated points}\;
$\cF \leftarrow \text{top frontier of }\texttt{convex\_hull}(Q(\os,\ocA) \setminus Q^+)$\tcp*[f]{candidate mixtures}\;
$\cF_Q \leftarrow \cF\cap Q(\os,\ocA)$\;
\For{points $q = Q(\os,\oa)\in\cF_Q$ in clockwise order}{
\uIf{find two successive points $((q_c^1, q_r^1), (q_c^2, q_r^2))$ of $\cF_Q$ such that $q_c^1 \leq \beta < q_c^2$}{
$p \leftarrow (\beta - q_c^1) / (q_c^2 - q_c^1)$\;
\Return the mixture $(1-p)\delta(\oa-\oa^1) + p\delta(\oa-\oa^2)$\;
}}
\lElse{\Return $\delta(\oa - \argmax_{\oa} Q_r(\os,\oa))$\tcp*[f]{budget $\beta$ always respected}}
\caption{Convex hull policy $\pi_\text{hull}(\oa|\os; Q)$}
\label{algo:pi_hull}
\end{algorithm}

\begin{figure}[ht]
    \centering
    \includegraphics[width=0.7\linewidth]{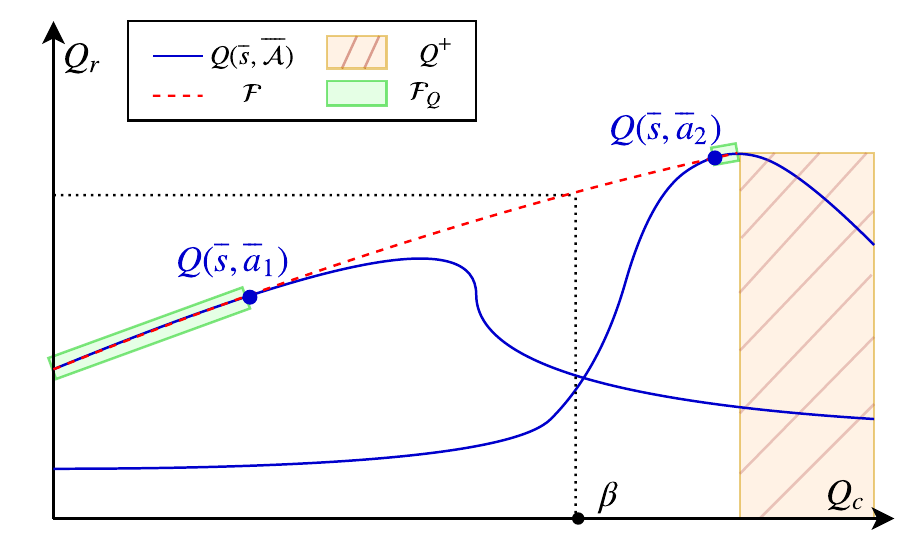}
    \caption{Representation of $\pi_\text{hull}$. When the budget lies between $Q(\os,\oa_1)$ and $Q(\os,\oa_2)$, two points of the top frontier of the convex hull, then the policy is a mixture of these two points.}
    \label{fig:hull}
\end{figure}

The computation of $\pi_\text{hull}$ in \Cref{algo:pi_hull} is illustrated in \Cref{fig:hull}.

\subsection{Function approximation}

Neural networks are well suited to model Q-functions in Reinforcement Learning algorithms \citep{Riedmiller2005,Mnih2015}.  We approximate $Q = (Q_r, Q_c)$ using one single neural network. Thus, the two components are jointly optimised which accelerates convergence and fosters learning of useful shared representations. Moreover, as in \citep{Mnih2015} we are dealing with a finite (categorical) action space $\cA$, instead of including the action in the input we add the output of the $Q$-function for each action to the last layer. Again, it provides a faster convergence toward useful shared representations and it only requires one forward pass to evaluate all action values. Finally, beside the state $s$ there is one more input to a budgeted $Q$-function:~the budget $\beta_a$. This budget is a scalar value whereas the state $s$ is a vector of potentially large size. To avoid a weak influence of $\beta$ compared to $s$ in the prediction, we include an additional encoder for the budget, whose width and depth may depend on the application. A straightforward choice is a single layer with the same width as the state. The overall architecture is shown in \Cref{fig:architecture} in \Cref{sec:bftq-full}.

\subsection{Parallel computing}
\label{subsec:parallel-computing}
In a simulated environment, a first process that can be distributed is the collection of samples in the exploration procedure of \Cref{algo:risk-sensitive-exploration}, as $\pi_\text{greedy}$ stays constant within each mini-batch which avoids the need of synchronisation between workers. Second, the main bottleneck of \BFTQ is the computation of the target $\cT Q$. Indeed, when computing $\pi_\text{hull}$ we must perform at each epoch a Graham-scan of complexity $\cO(|\cA||\tilde{\cB}| \log |\cA\tilde{\cB}|)$ per sample in $\mathcal{D}$ to compute the convex hulls of $Q$ (where $\tilde{\cB}$ is a finite discretisation of $\cB$). The resulting total time-complexity is $\cO(\frac{|\mathcal{D}||\cA||\tilde{\cB}|}{1-\gamma} \log |\cA||\tilde{\cB}|)$. This operation can easily be distributed over several CPUs provided that we first evaluate the model $Q(s',\cA\tilde{\cB})$ for each sample $s'\in\cD$, which can be done in a single forward pass. By using multiprocessing in the computations of $\pi_\text{hull}$, we enjoy a linear speedup.
The full description of our scalable implementation of \BFTQ is recalled in \Cref{algo:bftq_full} in \Cref{sec:bftq-full}.

\section{Experiments}
\label{sec:experiements}
There are two hypotheses we want to validate.

\paragraph{Exploration strategies}\label{par:ex-explo} We claimed in \Cref{sec:exploration} that a risk-sensitive exploration was required in the setting of BMDPs. We test this hypotheses by confronting our strategy to a classical risk-neutral strategy. The latter is chosen to be a $\epsilon$-greedy policy slowly transitioning from a random to a greedy policy\footnote{We train this greedy policy using \FTQ.} that aims to maximise $\expectedvalue_{\pi} G_r^\pi$ regardless of $\expectedvalue_{\pi} G_c^\pi$. The quality of the resulting batches $\cD$ is assessed by training a \BFTQ policy and comparing the resulting performance.

\paragraph{Budgeted algorithms}\label{par:ex-brl} We compare our  scalable \BFTQ algorithm described in \Cref{sec:scalable-bftq} to an \FTQl baseline. This baseline consists in approximating the BMDP by a finite set of CMDPs problems. We solve each of these CMDP using the standard technique of Lagrangian Relaxation: the cost constraint is converted to a soft penalty weighted by a Lagrangian multiplier $\lambda$ in a surrogate reward function: $\max_{\pi} \expectedvalue_{\pi}[G_r^\pi - \lambda G_c^\pi]$. The resulting MDP can be solved by any RL algorithm, and we chose \FTQ for being closest to \BFTQ.
In our experiments, a single training of \BFTQ corresponds to 10 trainings of \FTQl policies. Each run was repeated $N_{\text{seeds}}$ times. Parameters of the algorithms can be found in \Cref{sec:algorithms-parameters}

\subsection{Environments}
\label{subsec:environments}
We evaluate our method on three different environments involving reward-cost trade-offs. Their parameters can be found in \Cref{sec:env-parameters}

\paragraph{Corridors}
This simple environment is only meant to highlight clearly the specificity of exploration in a budgeted setting. It is a continuous gridworld with Gaussian perturbations, consisting in a maze composed of two corridors: a risky one with high rewards and costs, and a safe one with low rewards and no cost. In both corridors the outermost cell is the one yielding the most reward, which motivates a deep exploration.

\paragraph{Spoken dialogue system}
Our second application is a dialogue-based slot-filling simulation that has already benefited from batch RL optimisation in the past~\citep{Li2009ReinforcementLF,chandramohan2010optimizing,pietquin2011sample}. The system fills in a form of slot-values by interacting a user through speech, before sending them a response. For example, in a restaurant reservation domain, it may ask for three slots: the area of the restaurant, the price-range and the food type. The user could respectively provide those three slot-values : \texttt{Cambridge}, \texttt{Cheap} and \texttt{Indian-food}. In this application, we do not focus on how to extract such information from the user utterances, we rather focus on decision-making for filling in the form. To that end, the system can choose among a set of generic actions. As in \citep{carrara2018safe}, there are two ways of asking for a slot value: a slot value can be either be provided with an utterance, which may cause speech recognition errors with some probability, or by requiring the user to fill-in the slots by using a numeric pad. In this case, there are no recognition errors but a counterpart risk of hang-up: we assume that manually filling a key-value form is time-consuming and annoying. The environment yields a reward if all slots are filled without errors, and a constraint if the user hang-ups. Thus, there is a clear trade-off between using utterances and potentially committing a mistake, or using the numeric pad and risking a premature hang-up.

\paragraph{Autonomous driving}
In our third application, we use the \href{https://github.com/eleurent/highway-env}{highway-env} environment \citep{Leurent2018} for simulated highway driving and behavioural decision-making.
We define a task that displays a clear trade-off between safety and efficiency. The agent controls a vehicle with a finite set of manoeuvres implemented by low-lever controllers: $\mathcal{A}$ = \{\text{no-op}, \text{right-lane}, \text{left-lane}, \text{faster}, \text{slower}\}. It is driving on a two-lane road populated with other traffic participants: the vehicles in front of the agent drive slowly, and there are incoming vehicles on the opposite lane. Their behaviours are randomised, which introduces some uncertainty with respect to their possible future trajectories.
The task consists in driving as fast as possible, which is modelled by a reward proportional to the velocity: $R_r(s_t, a_t) \propto v_t$. This motivates the agent to try and overtake its preceding vehicles by driving fast on the opposite lane. This optimal but overly aggressive behaviour can be tempered through a cost function that embodies a safety objective: $R_c(s_t, a_t)$ is set to $1/H$ whenever the ego-vehicle is driving on the opposite lane, where $H$ is the episode horizon. Thus, the constrained signal $G_c^\pi$ is the maximum proportion of time that the agent is allowed to drive on the wrong side of the road.

\subsection{Results}
\label{subsec:results}
In the following figures, each patch represents the mean and 95\% confidence interval over $N_{\text{seeds}}$ seeds of the means of $(G_r^\pi,G_c^\pi)$ over $N_\text{trajs}$ trajectories. That way, we display the variation related to learning (and batches) rather than the variation in the execution of the policies.

\begin{figure}[tp]
    \centering
    \includegraphics[width=0.23\textwidth]{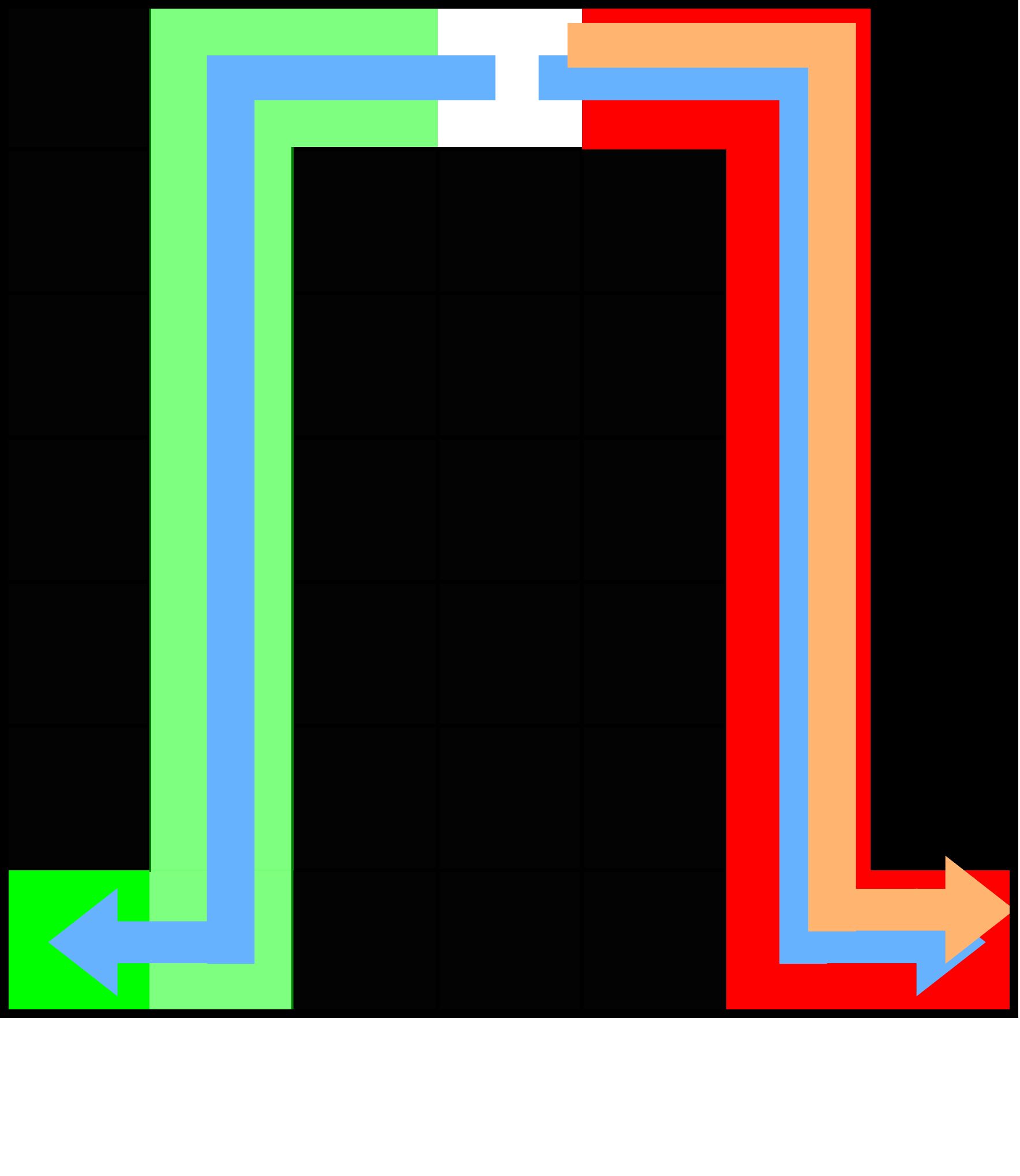}
    \includegraphics[page=1, width=0.45\textwidth]{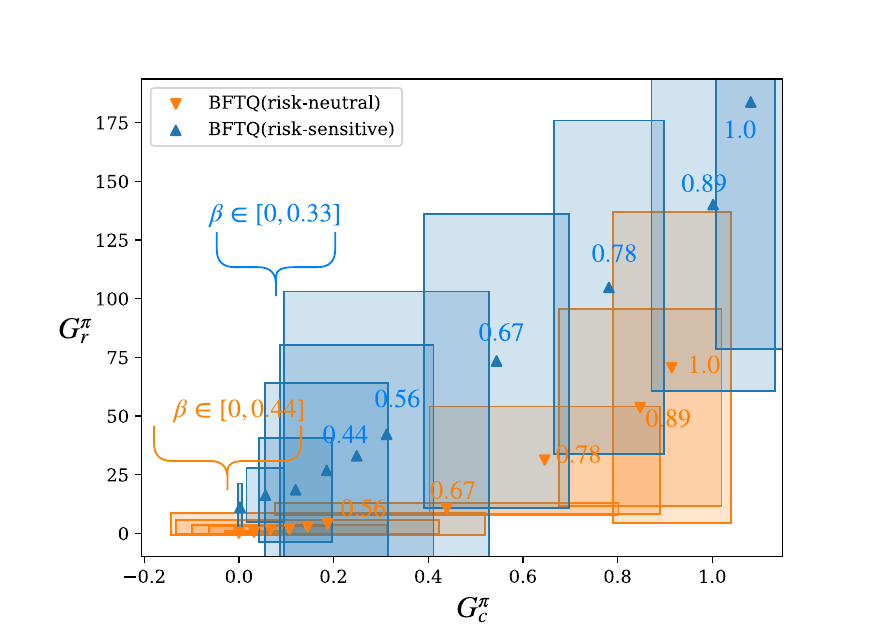}
    \caption{Trajectories (left) and performances (right) of two exploration strategies in the \texttt{corridors} environment. }
    \label{fig:exploration}
\end{figure}

We first bring to light the role of risk-sensitive exploration in the \text{corridors} environment: \Cref{fig:exploration} shows the set of trajectories collected by each exploration strategy\footnote{Animations are available in \Cref{subsec:exploration-examples}}, and the resulting performance of a budgeted policy trained on each batch. The trajectories (orange) in the risk-neutral batch are concentrated along the risky corridor (red) and ignore the safe corridor (green), which results in bad performances in the low-risk regime. Conversely, trajectories in the risk-sensitive batch (blue) are well distributed among both corridors and the corresponding budgeted policy achieves good performance across the whole spectrum of risk budgets.

\begin{figure}[tp]
    \begin{center}
    \includegraphics[width=0.49\linewidth]{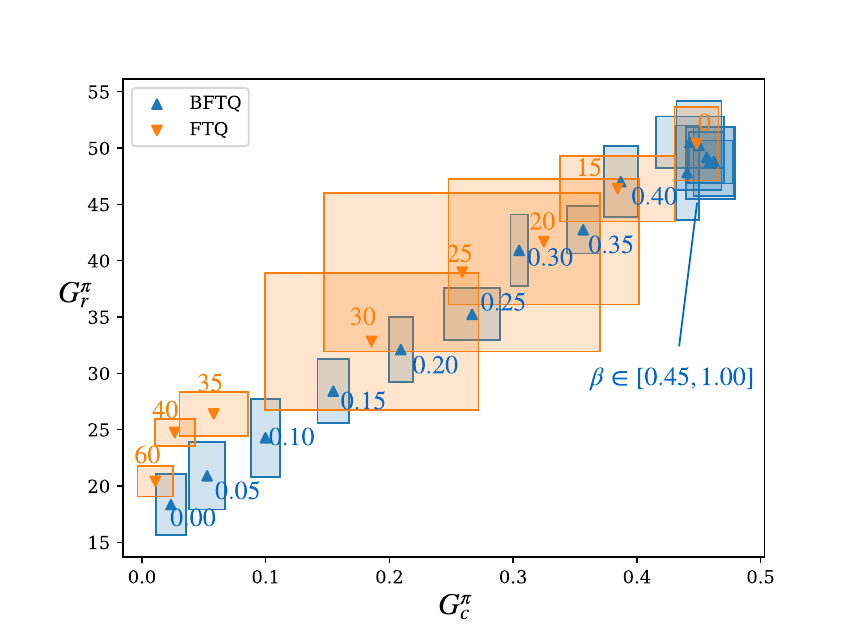}
    \includegraphics[width=0.49\linewidth]{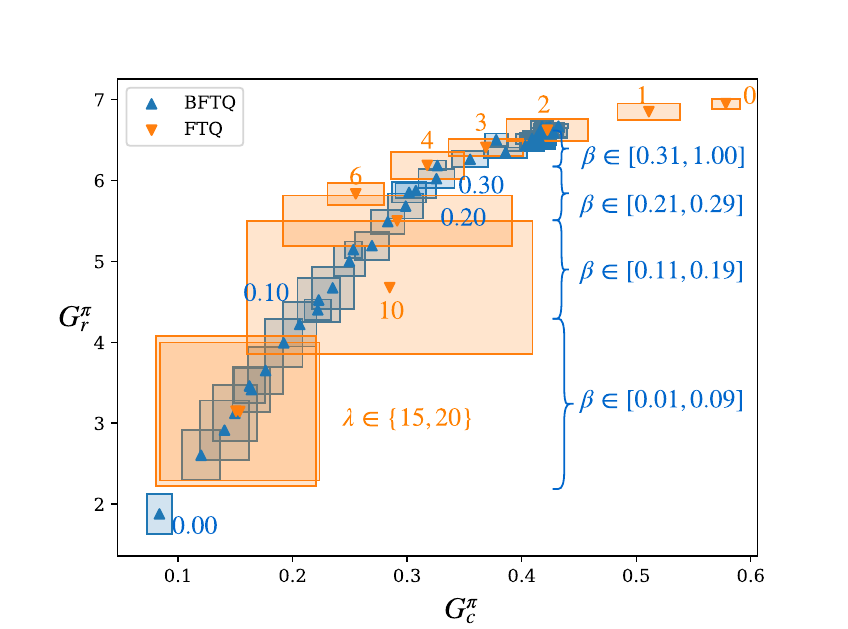}
    \caption{Performance comparison of \FTQl and \BFTQ on \text{slot-filling} (left) and \text{highway-env}(right) }
    \label{fig:results}
    \end{center}
\end{figure}

In a second experiment displayed in \Cref{fig:results}, we compare the performance of \FTQl to that of \BFTQ in the dialogue and autonomous driving tasks. For each algorithm, we plot the reward-cost trade-off curve. In both cases, \BFTQ performs almost as well as \FTQl despite only requiring a single model. All budgets are well-respected on \text{slot-filling}, but on \text{highway-env} we can observe an underestimation of $Q_c$, since e.g. $\expectedvalue[G_c|\beta=0] \simeq 0.1 $. This underestimation can be a consequence of two approximations: the use of the sampling operator $\hat{\cT}$ instead of the true environmental operator $\cT$, and the use of the neural network function approximation $Q_\theta$ instead of $Q$. 
Still, \BFTQ provides a better control on the expected cost of the policy, than \FTQl. In addition, \BFTQ behaves more consistently than \FTQl overall, as shown by its lower extra-seed variance. Examples of policy execution can be found in \Cref{sec:bftq-executions}.

\section{Discussion}
\label{subsec:discussions}
\Cref{algo:bftq} is an algorithm for solving large unknown BMDPs with continuous states. To the best of our knowledge, there is no algorithm in the current literature that combines all those features.


Algorithms have been proposed for CMDPs, which are less flexible sub-problems of the more general BMDP. When the environment parameters ($P$, $R_r$, $R_c$) are known but not tractable, solutions relying on function approximation~\citep{Undurti} or approximate linear programming~\citep{Poupart2015} have been proposed. For unknown environments, online algorithms \citep{Geibel2005, Abe2010,ChowGJP15,AchiamHTA17} and a batch algorithm \citep{Thomas2015, Petrik2016, Laroche2019,le2019batch} can solve large unknown CMDPs. Nevertheless, these approaches are limited in that the constraints thresholds are fixed prior to training and cannot be updated in real-time at policy execution to select the desired level of risk.

To our knowledge, there were only two ways of solving a BMDP. The first one is to approximate it with a finite set of CMDPs (e.g. see our \FTQl baseline). The solutions of these CMDPs take the form of mixtures between two deterministic policies \citep[Theorem 4.4,][]{BEUTLER1985236}. To obtain these policies, one needs to evaluate their expected cost by interacting with the environment\footnote{More details are provided in \Cref{sec:lagragian}}. Our solution not only requires one single model but also avoids any supplementary interaction.

The only other existing BMDP algorithm, and closest work to ours, is the Dynamic Programming algorithm proposed by \citet{Boutilier_Lu:uai16}. However, their work was established for finite state spaces only, and their solution relies heavily on this property. For instance, they enumerate and sort the next states $s'\in\cS$ by their expected value-by-cost, which could not be performed in a continuous state space $\cS$. Moreover, they rely on the knowledge of the model ($P$, $R_r$, $R_c$), and do not address the question of learning from interaction data.

\section{Conclusion}
\label{sec:conclusion}
The BMDP framework is a principled framework for safe decision making under uncertainty, which could be beneficial to the diffusion of Reinforcement Learning in industrial applications. However, BMDPs could so far only be solved in finite state spaces which limits their interest in many use-cases. We extend their definition to continuous states by introducing of a novel Dynamic Programming operator, that we build upon to propose a Reinforcement Learning algorithm. In order to scale to large problems, we provide an efficient implementation that exploits the structure of the value function and leverages tools from Deep Distributed Reinforcement Learning. We show that on two practical tasks our solution performs similarly to a baseline Lagrangian relaxation method while only requiring a single model to train, and relying on an interpretable $\beta$ instead of the tedious tuning of the penalty $\lambda$.

    \section*{Acknowledgments}
    This work has been supported by CPER Nord-Pas de Calais/FEDER DATA Advanced data science and technologies 2015-2020, the French Ministry of Higher Education and Research, INRIA, and the French Agence Nationale de la Recherche (ANR). We thank Guillaume Gautier, Fabrice Clerot, Xuedong Shang for the helpful discussions and valuable insights.
\bibliographystyle{named}
\bibliography{budgeted_rl}

\begin{thebibliography}{}

\bibitem[\protect\citeauthoryear{Abe and others}{2010}]{Abe2010}
Naoki Abe et~al.
\newblock Optimizing debt collections using constrained reinforcement learning.
\newblock In {\em Special Interest Group on Knowledge Discovery and Data Mining
  (SIGKDD)}, 2010.

\bibitem[\protect\citeauthoryear{Achiam \bgroup \em et al.\egroup
  }{2017}]{AchiamHTA17}
Joshua Achiam, David Held, Aviv Tamar, and Pieter Abbeel.
\newblock Constrained policy optimization.
\newblock In {\em Proceedings of the International Conference on Machine
  Learning (ICML)}, 2017.

\bibitem[\protect\citeauthoryear{Altman}{1999}]{Altman95constrainedmarkov}
Eitan Altman.
\newblock {\em Constrained Markov Decision Processes}.
\newblock CRC Press, 1999.

\bibitem[\protect\citeauthoryear{Beutler and Ross}{1985}]{BEUTLER1985236}
Frederick~J. Beutler and Keith~W. Ross.
\newblock Optimal policies for controlled markov chains with a constraint.
\newblock In {\em Journal of Mathematical Analysis and Applications}, 1985.

\bibitem[\protect\citeauthoryear{Boutilier and Lu}{2016}]{Boutilier_Lu:uai16}
Craig Boutilier and Tyler Lu.
\newblock Budget allocation using weakly coupled, constrained markov decision
  processes.
\newblock In {\em Uncertainty in Artificial Intelligence (UAI)}, 2016.

\bibitem[\protect\citeauthoryear{Carrara \bgroup \em et al.\egroup
  }{2018}]{carrara2018safe}
Nicolas Carrara, Romain Laroche, Jean-Léon Bouraoui, Tanguy Urvoy, and Olivier
  Pietquin.
\newblock Safe transfer learning for dialogue applications.
\newblock In {\em International Conference on Statistical Language and Speech
  Processing (SLSP)}, 2018.

\bibitem[\protect\citeauthoryear{Chandramohan \bgroup \em et al.\egroup
  }{2010}]{chandramohan2010optimizing}
Senthilkumar Chandramohan, Matthieu Geist, and Olivier Pietquin.
\newblock Optimizing spoken dialogue management with fitted value iteration.
\newblock In {\em Conference of the International Speech Communication
  Association (InterSpeech)}, 2010.

\bibitem[\protect\citeauthoryear{Chow \bgroup \em et al.\egroup
  }{2015}]{Chow2014}
Yinlam Chow, Aviv Tamar, Shie Mannor, and Marco Pavone.
\newblock {Risk-Sensitive and Robust Decision-Making: a CVaR Optimization
  Approach}.
\newblock In {\em Advances in Neural Information Processing Systems (NIPS)},
  2015.

\bibitem[\protect\citeauthoryear{Chow \bgroup \em et al.\egroup
  }{2018}]{ChowGJP15}
Yinlam Chow, Mohammad Ghavamzadeh, Lucas Janson, and Marco Pavone.
\newblock Risk-constrained reinforcement learning with percentile risk
  criteria.
\newblock In {\em Journal of Machine Learning Research (JMLR)}, 2018.

\bibitem[\protect\citeauthoryear{Dann \bgroup \em et al.\egroup
  }{2019}]{Dann2018}
Christoph Dann, Lihong Li, Wei Wei, and Emma Brunskill.
\newblock Policy certificates: Towards accountable reinforcement learning.
\newblock In {\em Proceedings of the International Conference on Machine
  Learning (ICML)}, 2019.

\bibitem[\protect\citeauthoryear{Ernst \bgroup \em et al.\egroup
  }{2005}]{Ernst2005}
Damien Ernst, Pierre Geurts, and Louis Wehenkel.
\newblock {Tree-Based Batch Mode Reinforcement Learning}.
\newblock In {\em Journal of Machine Learning Research (JMLR)}, 2005.

\bibitem[\protect\citeauthoryear{Garc{\'{i}}a and
  Fern{\'{a}}ndez}{2015}]{Garcia2015}
Javier Garc{\'{i}}a and Fernando Fern{\'{a}}ndez.
\newblock { A Comprehensive Survey on Safe Reinforcement Learning }.
\newblock In {\em Journal of Machine Learning Research (JMLR)}, 2015.

\bibitem[\protect\citeauthoryear{Geibel and Wysotzki}{2005}]{Geibel2005}
Peter Geibel and Fritz Wysotzki.
\newblock Risk-sensitive reinforcement learning applied to control under
  constraints.
\newblock In {\em Journal of Artificial Intelligence Research (JAIR)}, 2005.

\bibitem[\protect\citeauthoryear{Iyengar}{2005}]{Iyengar2005}
Garud~N. Iyengar.
\newblock { Robust Dynamic Programming }.
\newblock In {\em Mathematics of Operations Research}, 2005.

\bibitem[\protect\citeauthoryear{Khouzaimi \bgroup \em et al.\egroup
  }{2015}]{Khouzaimi2015}
Hatim Khouzaimi, Romain Laroche, and Fabrice. Lefevre.
\newblock { Optimising turn-taking strategies with reinforcement learning. }.
\newblock In {\em Special Interest Group on Discourse and Dialogue (SIGDIAL)},
  2015.

\bibitem[\protect\citeauthoryear{Laroche and Trichelair}{2019}]{Laroche2019}
Romain Laroche and R\'emi Trichelair, Paul and Tachet des~Combes.
\newblock Safe policy improvement with baseline bootstrapping.
\newblock In {\em Proceedings of the International Conference on Machine
  Learning (ICML)}, 2019.

\bibitem[\protect\citeauthoryear{Le \bgroup \em et al.\egroup
  }{2019}]{le2019batch}
Hoang~M. Le, Cameron Voloshin, and Yisong Yue.
\newblock Batch policy learning under constraints.
\newblock In {\em Proceedings of the International Conference on Machine
  Learning (ICML)}, 2019.

\bibitem[\protect\citeauthoryear{Leurent \bgroup \em et al.\egroup
  }{2018}]{Leurent2018}
Edouard Leurent, Yann Blanco, Denis Efimov, and Odalric-Ambrym Maillard.
\newblock { Approximate Robust Control of Uncertain Dynamical Systems }.
\newblock In {\em Neural Information Processing Systems (NeurIPS), Workshop on
  Machine Learning for Intelligent Transportation Systems}, 2018.

\bibitem[\protect\citeauthoryear{Li \bgroup \em et al.\egroup
  }{2009}]{Li2009ReinforcementLF}
Lihong Li, Jason~D. Williams, and Suhrid Balakrishnan.
\newblock Reinforcement learning for dialog management using least-squares
  policy iteration and fast feature selection.
\newblock In {\em Conference of the International Speech Communication
  Association (InterSpeech)}, 2009.

\bibitem[\protect\citeauthoryear{Liu \bgroup \em et al.\egroup
  }{2014}]{Liu2014}
Chunming Liu, Xin Xu, and Dewen Hu.
\newblock {Multiobjective Reinforcement Learning: A Comprehensive Overview}.
\newblock In {\em IEEE Transactions on Systems, Man, and Cybernetics: Systems},
  2014.

\bibitem[\protect\citeauthoryear{Luenberger}{2013}]{Luenberger2013}
David~G. Luenberger.
\newblock {\em Investment science}.
\newblock Oxford University Press, Incorporated, 2013.

\bibitem[\protect\citeauthoryear{Mausser and Rosen}{2003}]{Mausser2003}
H.~Mausser and D.~Rosen.
\newblock {Beyond VaR: from measuring risk to managing risk}.
\newblock In {\em Proceedings of the IEEE Conference on Computational
  Intelligence for Financial Engineering}, 2003.

\bibitem[\protect\citeauthoryear{Mnih \bgroup \em et al.\egroup
  }{2015}]{Mnih2015}
Volodymyr Mnih, Koray Kavukcuoglu, David Silver, Andrei~A. Rusu, Joel Veness,
  Marc~G. Bellemare, Alex Graves, Martin Riedmiller, Andreas~K. Fidjeland,
  Georg Ostrovski, Stig Petersen, Charles Beattie, Amir Sadik, Ioannis
  Antonoglou, Helen King, Dharshan Kumaran, Daan Wierstra, Shane Legg, and
  Demis Hassabis.
\newblock {Human-level control through deep reinforcement learning}.
\newblock {\em Nature}, 2015.

\bibitem[\protect\citeauthoryear{Nilim and {El Ghaoui}}{2005}]{Nilim2005}
Arnab Nilim and Laurent {El Ghaoui}.
\newblock { Robust Control of Markov Decision Processes with Uncertain
  Transition Matrices }.
\newblock In {\em Operations Research}, 2005.

\bibitem[\protect\citeauthoryear{Petrik \bgroup \em et al.\egroup
  }{2016}]{Petrik2016}
Mohammad Petrik, Marek~Ghavamzadeh, , and Yinlam Chow.
\newblock Safe policy improvement by minimizing robust baseline regret.
\newblock In {\em Advances in Neural Information Processing Systems (NIPS)},
  2016.

\bibitem[\protect\citeauthoryear{Pietquin \bgroup \em et al.\egroup
  }{2011}]{pietquin2011sample}
Olivier Pietquin, Matthieu Geist, Senthilkumar Chandramohan, and Herv{\'e}
  Frezza-Buet.
\newblock Sample-efficient batch reinforcement learning for dialogue management
  optimization.
\newblock {\em ACM Transactions on Speech and Language Processing (TSLP)},
  7(3):7, 2011.

\bibitem[\protect\citeauthoryear{Poupart \bgroup \em et al.\egroup
  }{2015}]{Poupart2015}
Pascal Poupart, Aarti Malhotra, Pei Pei, Kee-Eung Kim, Bongseok Goh, and
  Michael Bowling.
\newblock Approximate linear programming for constrained partially observable
  markov decision processes.
\newblock In {\em Proceedings of the Association for the Advancement of
  Artificial Intelligence Conference (AAAI)}, 2015.

\bibitem[\protect\citeauthoryear{Riedmiller}{2005}]{Riedmiller2005}
Martin Riedmiller.
\newblock {Neural fitted Q iteration - First experiences with a data efficient
  neural Reinforcement Learning method}.
\newblock In {\em Lecture Notes in Computer Science (including subseries
  Lecture Notes in Artificial Intelligence and Lecture Notes in
  Bioinformatics)}, 2005.

\bibitem[\protect\citeauthoryear{Roijers \bgroup \em et al.\egroup
  }{2013}]{Roijers2013ASO}
Diederik~M. Roijers, Peter Vamplew, Shimon Whiteson, and Richard Dazeley.
\newblock A survey of multi-objective sequential decision-making.
\newblock In {\em Journal of Artificial Intelligence Research (JAIR)}, 2013.

\bibitem[\protect\citeauthoryear{Tamar \bgroup \em et al.\egroup
  }{2012}]{Tamar2012}
Aviv Tamar, Dotan { Di Castro }, and Shie Mannor.
\newblock { Policy Gradients with Variance Related Risk Criteria }.
\newblock In {\em Proceedings of the International Conference on Machine
  Learning (ICML)}, 2012.

\bibitem[\protect\citeauthoryear{Thomas \bgroup \em et al.\egroup
  }{2015}]{Thomas2015}
Philip Thomas, Georgios Theocharous, and Mohammad Ghavamzadeh.
\newblock High confidence policy improvement.
\newblock In {\em Proceedings of the International Conference on Machine
  Learning (ICML)}, 2015.

\bibitem[\protect\citeauthoryear{Undurti \bgroup \em et al.\egroup
  }{2011}]{Undurti}
Aditya Undurti, Alborz Geramifard, and Jonathan~P. How.
\newblock Function approximation for continuous constrained mdps.
\newblock In {\em Tech Report}, 2011.

\bibitem[\protect\citeauthoryear{Wiesemann \bgroup \em et al.\egroup
  }{2013}]{Wiesemann2013}
Wolfram Wiesemann, Daniel Kuhn, and Berç Rustem.
\newblock Robust markov decision processes.
\newblock In {\em Mathematics of Operations Research}, 2013.

\end{thebibliography}

\clearpage
\begin{center}
\LARGE Appendices
\end{center}
\appendix

\paragraph{Outline}

This paper gathers all the supplementary material and goes as follows: \Cref{sec:proofs} details all the proofs of the main results. \Cref{sec:risk-sensitive-supp} and \Cref{sec:bftq-full} recall respectively the scalable \BFTQ algorithm and the risk-sensitive exploration procedure. \Cref{sec:lagragian} describes a naive alternative to \BFTQ based on Lagrangian Relaxation. The \Cref{sec:exp-supp} assembles all the assets for visualising and reproducing the experiments, including visualisations of policy executions, algorithms and environment parameters, and instructions for executing the attached source code. Finally we fill the Machine Learning Reproducibility Checklist and we justify each statement in \Cref{sec:ml-checklist}.

\section{Proofs of Main Results}
\label{sec:proofs}
\subsection{\Cref{prop:bellman-expectation}}

\begin{proof}
This proof is the same as that in classical multi-objective MDPs.

\begin{align*}
    V^\pi(\os) &\eqdef \expectedvalue\left[ G^\pi \condbar \ov{s_0} = \os\right] \\
    &=\sum_{\oa\in\ocA} \probability{\oa_0 = \oa \condbar\ov{s_0} = \os} \expectedvalue\left[ G^\pi \condbar \ov{s_0} = \os, \oa_0 = \oa\right]\\
    &= \sum_{\oa\in\ocA} \pi(\oa | \os) Q^\pi(\os,\oa)
\end{align*}
\begin{align*}
    Q^\pi(\os, \oa) &\eqdef \expectedvalue\left[\sum_{t=0}^\infty \gamma^t R(\os_t, \oa_t)\condbar \ov{s_0} = \os, \ov{a_0} = \oa\right] \\
    &= R(\os, \oa) + \sum_{\os'\in\ocS}\probability{\os_1 = \os' \condbar\ov{s_0} = \os, \ov{a_0} = \oa}\cdot \expectedvalue\left[\sum_{t=1}^\infty \gamma^t R(\os_t, \oa_t)\condbar \ov{s_1} = \os'\right] \\
    &= R(\os, \oa) + \gamma\sum_{\os'\in\ocS}\ov{P}\left(\os' \condbar\os, \oa\right) \expectedvalue\left[\sum_{t=0}^\infty \gamma^t R(\os_t, \oa_t) \condbar \ov{s_0} = \os'\right] \\
    &=  R(\os, \oa) + \gamma\sum_{\os'\in\ocS}\ov{P}\left(\os' \condbar\os, \oa\right) V^\pi(\os')
\end{align*}

\textbf{Contraction of $\cT^\pi$:}
Let $\pi\in\Pi, Q_1, Q_2\in(\Real^2)^{\ocS\ocA}$.
\begin{align*}
    \forall \os\in\ocS, \oa\in\ocA,\quad \left|\cT^\pi Q_1(\os,\oa) - \cT^\pi Q_2(\os,\oa)\right| &= \left|\gamma\expectedvalueover{\substack{\os'\sim\ov{P}(\os'|\os,\oa) \\ \oa'\sim\pi(\oa'|\os')}} Q_1(\os',\oa') - Q_2(\os',\oa')\right|\\
    &\leq \gamma\left\|Q_1-Q_2\right\|_\infty
\end{align*}
Hence, $\left\|\cT^\pi Q_1  - \cT^\pi Q_2 \right\|_\infty \leq \gamma\left\|Q_1-Q_2\right\|_\infty$

According to the Banach fixed point theorem, $\cT^\pi$ admits a unique fixed point.
It can be easily verified that $Q^\pi$ is indeed this fixed point by combining the two Bellman Expectation equations \eqref{eq:bellman_expectation}.

\end{proof}

\subsection{\Cref{thm:bellman-optimality}}

\begin{proof}
Let $\os, \oa \in \ocA\times\ocS$. For this proof, we consider  potentially non-stationary policies $\pi=(\rho, \pi')$, with $\rho\in\cM(\ocA)$, $\pi'\in\cM(\ocA)^\Natural$. The results will apply to the particular case of stationary optimal policies, when they exist.

\begin{align}
    Q_r^*(\os, \oa) &=  \max_{\rho, \pi'} Q_r^{\rho, \pi'}(\os', \oa') \label{eq:pthm_def}\\
    &= \max_{\rho, \pi'} R_r(\os, \oa) + \gamma \sum_{\os'\in\cS} P(\os' | \os, \oa) V_r^{\rho, \pi'}(\os') \label{eq:pthm_exp}\\
    &= R_r(\os, \oa) + \gamma \sum_{\os'\in\cS}  P(\os' | \os, \oa) \max_{\rho, \pi'} \sum_{\oa'\in\ocA} \rho(\oa' | \os')Q_r^{\pi'}(\os', \oa') \label{eq:pthm_marg}\\
    &= R_r(\os, \oa) + \gamma \sum_{\os'\in\cS}  P(\os' | \os, \oa) \max_\rho\sum_{\oa'\in\ocA}\rho(\oa' | \os')\max_{\pi'\in\Pi_a(\os')}Q_r^{\pi'}(\os', \oa') \label{eq:pthm_max}\\
    &= R_r(\os, \oa) + \gamma \sum_{\os'\in\cS}  P(\os' | \os, \oa) \max_\rho\expectedvalueover{\oa'\sim\rho}Q_r^*(\os', \oa') \label{eq:pthm_marg_def2}
\end{align}
where $\pi = (\rho, \pi')\in\Pi_a(\os)$ and $\pi'\in\Pi_a(\os')$.

This follows from:
\begin{enumerate}
\item[\eqref{eq:pthm_def}.] Definition of $Q^*$. 
\item[\eqref{eq:pthm_exp}.] Bellman Expectation expansion from \Cref{prop:bellman-expectation}.
\item[\eqref{eq:pthm_marg}.] Marginalisation on $\oa'$.
\item[\eqref{eq:pthm_max}.] \begin{itemize}
    \item Trivially $\max_{\pi'\in\Pi_a(\os')} \sum_{\oa'\in\cA} \cdot \leq \sum_{\oa'\in\cA} \max_{ \pi'\in\Pi_a(\os)} \cdot$
    \item Let $\ov{\pi}\in\argmax_{\pi'\in\Pi_a(\os')} Q_r^{\pi'}(\os', \oa')$, then:
    \begin{align*}
        \sum_{\oa'\in\ov{A}}\rho(\oa'|\os')\max_{\pi'\in\Pi_a(\os')}Q_r^{\pi'}(\os', \oa') &= \sum_{\oa'\in\ov{A}}\rho(\oa'|\os')Q_r^{\ov{\pi}}(\os', \oa') \\
        &\leq  \max_{\pi'\in\Pi_a(\os')} \sum_{\oa'\in\ov{A}}\rho(\oa'|\os')Q_r^{\pi'}(\os', \oa')
    \end{align*}
\end{itemize}
\item[\eqref{eq:pthm_marg_def2}.] Definition of $Q^*$.
\end{enumerate}

Moreover, the condition $\pi=(\rho, \pi')\in\Pi_a(\os)$ gives
\begin{equation*}
   \expectedvalueover{\oa'\sim\rho} Q_c^{*}(\os, \oa) = \expectedvalueover{\oa'\sim\rho} Q_c^{\pi'}(\os, \oa) = V_c^{\pi}(\os) \leq \beta
\end{equation*}

Consequently, $\pi_\text{greedy}(\cdot; Q^*)$ belongs to the $\argmax$ of \eqref{eq:pthm_marg_def2}, and in particular:
\begin{equation*}
     Q_r^*(\os, \oa) = r(\os, \oa) + \gamma \sum_{\os'\in\cS}  P(\os' | \os, \oa) \expectedvalueover{\oa'\sim\pi_\text{greedy}(\os', Q^*)} Q_r^*(\os', \oa')
\end{equation*}

The same reasoning can be made for $Q_c^*$ by replacing $\max$ operators by $\min$, and $\Pi_a$ by $\Pi_r$.
\end{proof}

\subsection{\Cref{prop:greedy_optimal}}
\begin{proof}
Notice from the definitions of $\cT$ and $\cT^\pi$ in \eqref{eq:bellman-optimality} and \eqref{eq:bellman_expectation_operator} that $\cT$ and $\cT^{\pi_\text{greedy}(\cdot;Q^*)}$ coincide on $Q^*$. Moreover, since $Q^* = \cT Q^*$ by \Cref{thm:bellman-optimality}, we have: $    \cT^{\pi_\text{greedy}(\cdot;Q^*)} Q^* = \cT Q^* = Q^*
$.
Hence, $Q^*$ is a fixed point of $\cT^{\pi_\text{greedy}(\cdot;Q^*)}$, and by \Cref{prop:bellman-expectation} it must be equal to $Q^{\pi_\text{greedy}(\cdot;Q^*)}$

To show the same result for $V^*$, notice that 
\begin{equation*}
    V^{\pi_\text{greedy}(Q^*)}(\os) = \expectedvalueover{\oa\sim\pi_\text{greedy}(Q^*)}Q^{\pi_\text{greedy}(Q^*)}(\os,\oa) = \expectedvalueover{\oa\sim\pi_\text{greedy}(Q^*)}Q^*(\os,\oa)
\end{equation*}
By applying the definitions of $Q^*$ and $\pi_\text{greedy}$, we recover the definition of $V^*$.
\end{proof}

\subsection{\Cref{thm:contraction}}
\label{sec:proof_contraction}
\begin{proof}
In the trivial case $|\cA| = 1$, there exits only one policy $\pi$ and $\cT = \cT^\pi$, which is a contraction by \Cref{prop:bellman-expectation}.

In the general case $|\cA| \geq 2$, we can build the following counter-example:

Let $(\cS, \cA, P, R_r, R_c)$ be a BMDP.
For any $\epsilon > 0$, we define $Q_\epsilon^1$ and $Q_\epsilon^2$ as:

\begin{align*}
      Q_\epsilon^1(\os,\oa) =
      \begin{cases}
    (0, 0), & \text{if } a = a_0 \\
    \left(\frac{1}{\gamma}, \epsilon\right), & \text{if } a \neq a_0
  \end{cases}\\
  Q_\epsilon^2(\os,\oa) =
      \begin{cases}
    (0, \epsilon), & \text{if } a = a_0 \\
    \left(\frac{1}{\gamma}, 2\epsilon\right), & \text{if } a \neq a_0
  \end{cases}
\end{align*}
Then, $\|Q_1-Q_2\|_\infty = \epsilon$.
 $Q_\epsilon^1$ and $Q_\epsilon^2$ are represented in \Cref{fig:concavity_example}.

\begin{figure}[tp]
    \centering
    \includegraphics[width=0.5\textwidth]{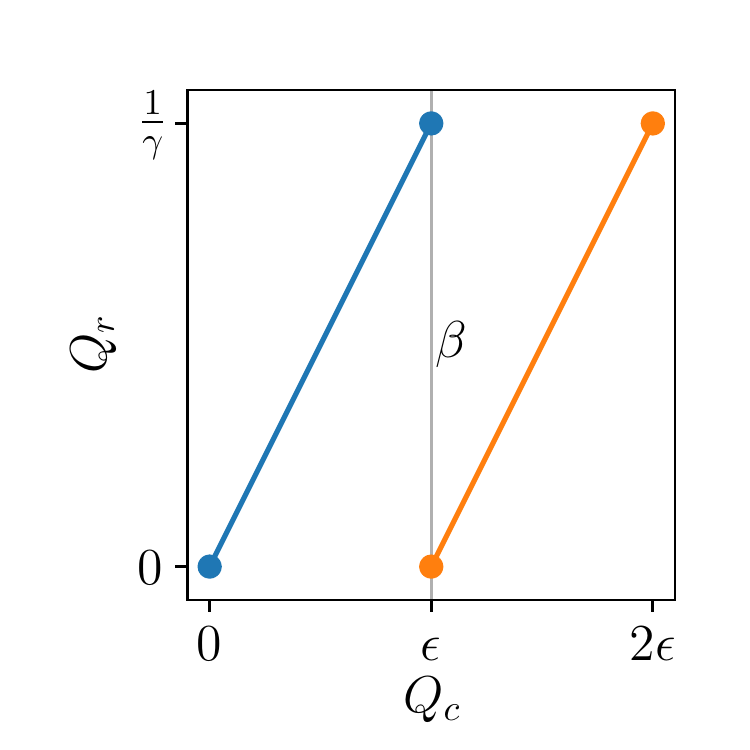}
    \caption{Representation of $Q_\epsilon^1$ (blue) and $Q_\epsilon^2$ (yellow)}
    \label{fig:concavity_example}
\end{figure}

But for $\oa=(a,\beta_a)$ with $\beta_a = \epsilon$, we have:
\begin{align*}
    \|\cT Q_\epsilon^1(\os, \oa) - \cT Q_\epsilon^2(\os, \oa)\|_\infty &= \gamma\left\|\expectedvalueover{\os'\sim\ov{P}(\os'|\os,\oa)} \expectedvalueover{\oa'\sim\pi_\text{greedy}(Q^1_\epsilon)}Q^1_\epsilon(\os',\oa') - \expectedvalueover{\oa'\sim\pi_\text{greedy}(Q^2_\epsilon)}Q^2_\epsilon(\os',\oa')\right\|_\infty \\
    &= \gamma\left\|\expectedvalueover{\os'\sim\ov{P}(\os'|\os,\oa)}\left(\frac{1}{\gamma}, \epsilon\right) - (0, \epsilon)\right\|_\infty \\
    &= \gamma\frac{1}{\gamma} = 1
\end{align*}
Hence, 
\begin{align*}
    \|\cT Q_\epsilon^1 - \cT Q_\epsilon^2\|_\infty &\geq 1 = \frac{1}{\epsilon} \|Q_1-Q_2\|_\infty
\end{align*}

In particular, there does not exist $L>0$ such that:
$$\forall Q_1,Q_2\in(\Real^2)^{\ocS\ocA}, \|\cT Q^1 - \cT Q^2\|_\infty \leq L \|Q^1 - Q^2\|_\infty$$
In other words, $\cT$ is not a contraction for $\|\cdot\|_\infty$.
\end{proof}

\subsection{\Cref{rmk:contractivity-smooth}}
\label{proof:contraction-with-smooth}
\begin{proof}
We now study the contractivity of $\cT$ when restricted to the functions of $\cL_\gamma$ defined as follows:
\begin{equation}
    \cL_\gamma = \left\{\begin{array}{cc}
         Q\in(\Real^2)^{\ocS\ocA}\text{ s.t. }\exists L<\frac{1}{\gamma}-1: \forall \os\in\ocS,\oa_1,\oa_2\in\ocA,   \\
         |Q_r(\os,\oa_1) - Q_r(\os,\oa_2)| \leq L|Q_c(\os,\oa_1) - Q_c(\os,\oa_2)|
    \end{array}\right\}
\end{equation}
That is, for all state $\os$, the set $Q(\os, \ocA)$ plot in the $(Q_c,Q_r)$ plane must be the \emph{graph} of a $L$-Lipschitz function, with $L<1/\gamma-1$.

We impose such structure for the following reason: the counter-example presented above prevented contraction because it was a pathological case in which the slope of $Q$ can be arbitrary large. As a consequence, when solving $Q_r^*$ such that $Q_c^*=\beta$, a vertical slice of a $\|\cdot\|_\infty$ ball around $Q_1$ (which must contain $Q_2$) can be arbitrary large as well.

This sketch of proof makes use of insights detailed in the proof of \Cref{prop:bftq_pi_hull}, which we recommend the reader to consult first.

We denote $\cB(Q,R)$ the ball of centre $Q$ and radius $R$ for the $\|\cdot\|_\infty$-norm:
\begin{equation*}
    \cB(Q,R) = \{Q'\in(R^2)^{\ocS\ocA}: \|Q-Q'\|_\infty \leq R\}
\end{equation*}

We give the three main steps required to show that $\cT$ restricted to $\cL_\gamma$ is a contraction. Given $Q^1, Q^2\in\cL_\gamma$, show that:
\begin{enumerate}
    \item $Q^2\in\cB(Q^1,R)\implies\cF^2\in\cB(\cF^1, R), \forall\os\in\ocS$, where $\cF$ is the top frontier of the convex hull of undominated points, as defined in \Cref{sec:proof_pi_hull}.
    \item $Q\in\cL_\gamma \implies \cF$ is the graph of a $L$-Lipschitz function, $\forall\os\in\ocS$.
    \item taking the slice $Q_c=\beta$ of a ball $\cB(\cF,R)$ with $\cF$ $L$-Lipschitz results in an interval on $Q_r$ of range at most $(L+1)R$
\end{enumerate}

These three steps will allow us to control $Q_r^{2*} - Q_r^{1*}$ as a function of $R = \|Q^2-Q^1\|_\infty$.

\textbf{Step 1:} we want to show that if $Q^1$ and $Q^2$ are close, then $\cF^1$ are $\cF^2$ are close as well in the following sense:
\begin{align}
    \cF^2\in\cB(\cF^1, R) &\iff d(\cF^1, \cF^2) \leq R \iff \max_{q^2\in\cF^2}\min_{q^1\in\cF^1}\|q^2-q^1\|_\infty \leq R
    \label{eq:ball-set}
\end{align}
Assume $Q^2\in\cB(Q^1,R)$.
We start by showing this result for $\cC^2(Q^{1-})$ and $\cC^2(Q^{2-})$ as defined in \Cref{sec:proof_pi_hull}:

Let $\os\in\ocS$ and $q^2\in\cC^2(Q^{2-})$, $\exists\lambda\in[0,1], \oa_1,\oa_2\in\ocA: q^2 = (1-\lambda)Q^2(\os,\oa_1) + \lambda Q^2(\os,\oa_2)$. Define $q^1 = (1-\lambda)Q^1(\os,\oa_1) + \lambda Q^1(\os,\oa_2)$. Then 
\begin{align*}
    \|q^2-q^1\|_\infty &= \|(1-\lambda)(Q^2(\os,\oa_1) - Q^1(\os,\oa_1)) + \lambda (Q^2(\os,\oa_2) - Q^1(\os,\oa_2))\|_\infty\\
    &\leq  (1-\lambda)\|Q^2(\os,\oa_1) - Q^1(\os,\oa_1)\|_\infty + \lambda \|Q^2(\os,\oa_2) - Q^1(\os,\oa_2)\|_\infty\\
    &\leq (1-\lambda)R+\lambda R = R
\end{align*}

It remains to show that when taking the top frontiers of the convex sets $\cC^2(Q^{1-})$ and $\cC^2(Q^{2-})$, they remain at a distance of at most $R$.

This is illustrated in \Cref{fig:contraction_lips_hull}: given a function $Q^1$, we show the locus $\cB(Q_1,R)$ of $Q^2$. We then draw $\cF^1$ the top frontier of the convex hull of $Q^1$ and alongside the locus of all possible $\cF^2$, which belong to a ball $\cB(\cF^1, R)$. 

\begin{figure}[ht]
    \centering
    \includegraphics[trim=7cm 4cm 7cm 4cm, clip, width=0.7\textwidth]{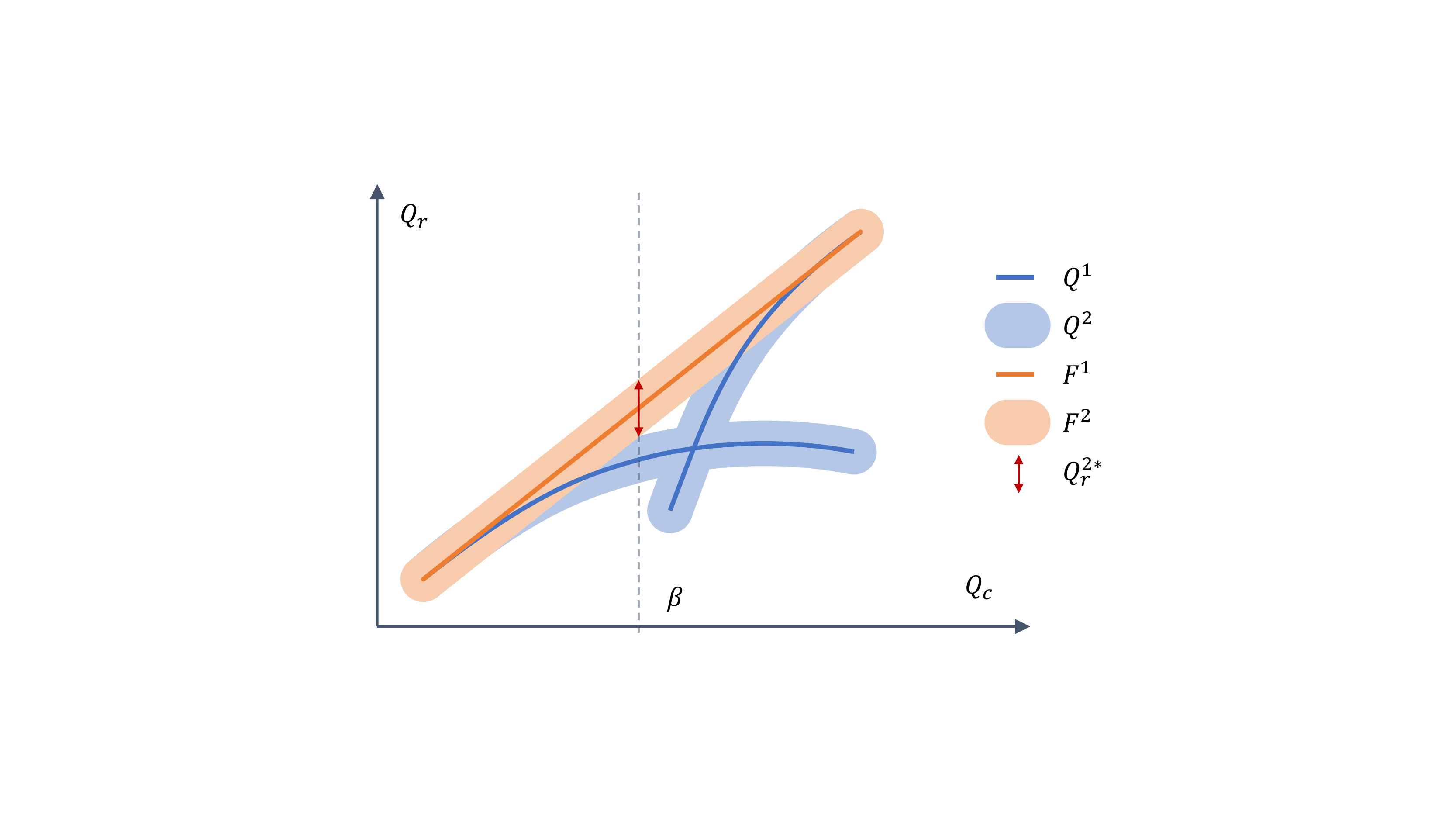}
    \caption{We represent the range of possible solutions $Q_r^{2,*}$ for any $Q^2\in\cB(Q^1)$, given $Q_1\in\cL_\lambda$}
    \label{fig:contraction_lips_hull}
\end{figure}

\textbf{Step 2:} We want to show that if $Q\in\cL_\gamma$, $\cF$ is the graph of an $L$-Lipschitz function:
\begin{equation}
\label{eq:L-lip-set}
    \forall q^1,q^2\in\cF, |q_r^2-q_r^1| \leq |q_c^2-q_c^1|
\end{equation}

Let $Q\in\cL_\gamma$ and $\os\in\ocS$, $\cF$ the corresponding top frontier of convex hull.
For all $q^1,q^2\in\cF, \exists \lambda,\mu\in[0,1], q^{11},q^{12},q^{21},q^{22}\in Q(\os,\ocA)$ such that $q^1 = (1-\lambda)q^{11} + \lambda q^{12}$ and $q^2 = (1-\mu)q^{21} + \mu q^{22}$.
Without loss of generality, we can assume $q_c^{11}\leq q_c^{12}$ and $q_c^{21}\leq q_c^{22}$. We also consider the worst case in terms of maximum $q_r$ deviation: $q_c^{12} \leq q_c^{21}$.
Then the maximum increment $q_r^2-q_r^{1}$ is:
\begin{align*}
    \|q^2_r-q^{1}_r\| &\leq \|q^{12}_r-q^{1}_r\| + \|q^{21}_r-q^{12}_r\| + \|q^{2}_r-q^{21}_r\| \\
    &= (1-\lambda)\|q^{12}_r-q^{11}_r\| + \|q^{21}_r-q^{12}_r\| + \mu\|q^{22}_r-q^{21}_r\| \\ 
    &\leq (1-\lambda)L\|q^{12}_c-q^{11}_c\| + L\|q^{21}_c-q^{12}_c\| + \mu L\|q^{22}_c-q^{21}_c\| \\
    &= L\|q^{12}_c-q^{1}_c\| + L\|q^{21}_c-q^{12}_c\| + L\|q^{2}_c-q^{21}_c\|\\
    &= L\|q^{2}_c-q^{1}_c\|
\end{align*}

This can also be seen in \Cref{fig:contraction_lips_hull}: the maximum slope of the $\cF^1$ is lower than the maximum slope between two points of $Q^1$.

\textbf{Step 3:} Let $\cF_1$ be a L-Lipschitz set as defined in \eqref{eq:L-lip-set}, and consider a ball $\cB(\cF_1,R)$ around it as defined in \eqref{eq:ball-set}.

We want to bound the optimal reward value $Q_r^{2*}$ under constraint $Q_c^{2*} = \beta$ (regular case in \Cref{sec:proof_pi_hull} where the constraint is saturated), for any $\cF^2\in\cB(\cF_1,R)$. This quantity is represented as a red double-ended arrow in \Cref{fig:contraction_lips_hull}.

Because we are only interested in what happens locally at $Q_c=\beta$, we can zoom in on \Cref{fig:contraction_lips_hull} and only consider a thin $\epsilon$-section around $\beta$. In the limit $\epsilon\rightarrow 0$, this section becomes the tangent to $\cF^1$ at $Q_c^1=\beta$. It is represented in \Cref{fig:contraction_lips_hull_slope}, from which we derive a geometrical proof:
\begin{figure}[ht]
    \centering
    \includegraphics[trim=2cm 1cm 2cm 1cm, clip, width=0.7\textwidth]{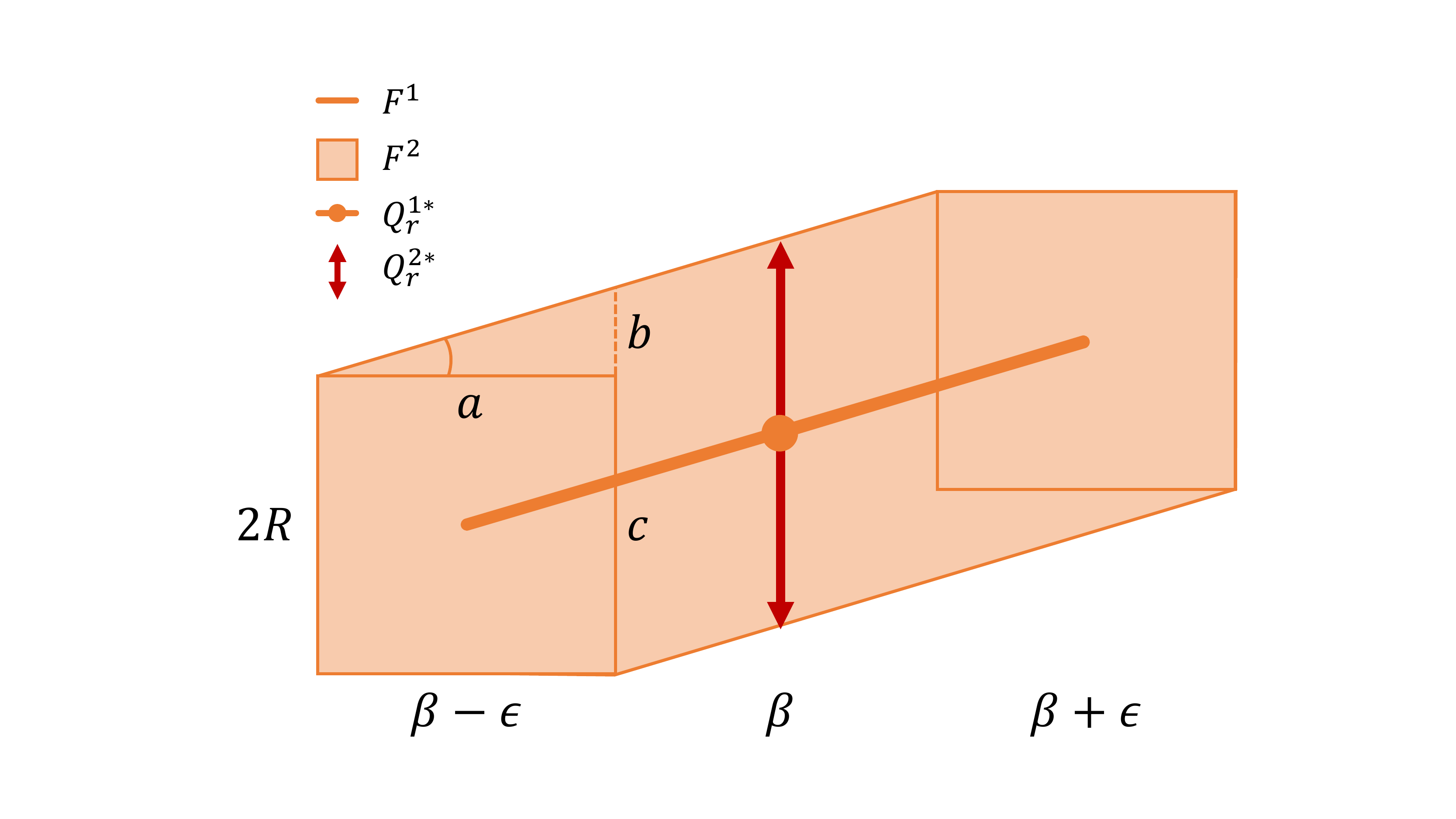}
    \caption{We represent a section $[\beta-\epsilon, \beta+\epsilon]$ of $\cF^1$ and $B(\cF^1, R)$. We want to bound the range of $Q_r^{2*}.$}
    \label{fig:contraction_lips_hull_slope}
\end{figure}

\begin{align*}
    \Delta Q_r^{2*} &= b + c &\\
    & \leq La + c & \text{($\cF^1$ $L$-Lipschitz)}\\
    &= 2LR+2R = 2R(L+1)
\end{align*}
Hence,
\begin{equation*}
    | Q_r^{2*} - Q_r^{1*}| \leq \frac{\Delta Q_r^{2*}}{2} = R(L+1)
\end{equation*}

and $Q_c^{1*} = Q_c^{2*} = \beta$.
Consequently, $ \|Q^{2*} - Q^{1*}\|_\infty \leq (L+1)R$

For completeness, the edge case in \Cref{sec:proof_pi_hull} should be considered as well.

\textbf{Wrapping it up:}

We've shown that for any $Q^1,Q^2\in\cL_\gamma$, and all $\os\in\ocS$, $\cF^2\in\cB(\cF^1,\|Q^2-Q^1\|_\infty)$ and $\cF^1$ is the graph of a $L$-Lipschitz function with $L<1/\gamma - 1$. Moreover, the solutions of $\pi_\text{greedy}(Q^1)$ and $\pi_\text{greedy}(Q^2)$ at $\os$ are such that $ \|Q^{2*} - Q^{1*}\|_\infty \leq (L+1)\|Q^2-Q^1\|_\infty$.

Hence, for all $\oa$,
\begin{align*}
    \|\cT Q^1(\os, \oa) - \cT Q^2(\os, \oa)\|_\infty &= \gamma\left\|\expectedvalueover{\os'\sim\ov{P}(\os'|\os,\oa)} \expectedvalueover{\oa'\sim\pi_\text{greedy}(Q^1)}Q^1(\os',\oa') - \expectedvalueover{\oa'\sim\pi_\text{greedy}(Q^2)}Q^2(\os',\oa')\right\|_\infty \\
    &= \gamma\left\|Q^{2*} - Q^{1*}\right\|_\infty \\
    &\leq \gamma(L+1)\|Q^2-Q^1\|_\infty
\end{align*}
Taking the sup on $\ocS\ocA$,

\begin{equation*}
    \|\cT Q^1 - \cT Q^2\|_\infty \leq \gamma(L+1)\|Q^1-Q^2\|_\infty
\end{equation*}
with $\gamma(L+1) < 1$.
As a conclusion, $\cT$ is a $\gamma(L+1)$-contraction on $\cL_\gamma$.
\end{proof}

\subsection{\Cref{prop:bftq_pi_hull}}
\label{sec:proof_pi_hull}
\begin{definition}
Let $A$ be a set, and $f$ a function defined on $A$. We define:

\begin{itemize}
    \item Convex hull of $A$: $\cC(A) = \{\sum_{i=1}^p \lambda_i a_i: a_i\in A, \lambda_i\in\Real^+, \sum_{i=1}^p \lambda_i = 1, p\in\Natural\}$
    \item Convex edges of $A$: $\cC^2(A) = \{\lambda a_1 + (1-\lambda)a_2: a_1, a_2\in A, \lambda\in[0, 1]\}$
    \item Dirac distributions of $A$: $\delta(A) = \{\delta(a-a_0): a_0\in A\}$ 
    \item Image of $A$ by $f$: $f(A) = \{f(a): a\in A\}$
\end{itemize}
\end{definition}

\begin{proof}
Let $\os=(s,\beta)\in\ocS$ and $Q\in(\Real^2)^{\ocS\ocA}$. We recall the definition of $\pi_\text{greedy}$:
\begin{subequations}
\begin{equation}
    \pi_\text{greedy}(\oa|\os; Q) \in \argmin_{\rho\in\Pi_r^Q} \expectedvalueover{\oa\sim\rho}Q_c(\os, \oa) \tag{\ref{eq:pi_greedy_cost}}
\end{equation}
\begin{align}
    \text{where }\quad\Pi_r^Q = &\argmax_{\rho\in\cM(\ocA)} \expectedvalueover{\oa\sim\rho} Q_r(\os, \oa) \tag{\ref{eq:pi_greedy_reward}}\\
    & \text{ s.t. }  \expectedvalueover{\oa\sim\rho} Q_c(\os, \oa) \leq \beta \tag{\ref{eq:pi_greedy_constraint}}
\end{align}
\end{subequations}

Note that any policy in the $\argmin$ in \eqref{eq:pi_greedy_cost} is suitable to compute $\cT$.
We first reduce the set of candidate optimal policies.
Consider the problem described in \eqref{eq:pi_greedy_reward},\eqref{eq:pi_greedy_constraint}: it can be seen as a single-step CMDP problem with reward $R_r=Q_r$ and cost $R_c=Q_c$. By \citep[Theorem 4.4][]{BEUTLER1985236}, we know that the solutions are mixtures of two deterministic policies. Hence, we can replace $\cM(\cA)$ by $\cC^2(\delta(\ocA))$ in \eqref{eq:pi_greedy_reward}.

Moreover, remark that:
\begin{align*}
    \{\expectedvalueover{\oa\sim\rho} Q(\os,\oa): \rho\in \cC^2(\delta(\ocA))\} &= \{\expectedvalueover{\oa\sim\rho} Q(\os,\oa): \rho=(1-\lambda)\delta(\oa-\oa_1)+\lambda\delta(\oa-\oa_2), \oa_1,\oa_2\in\ocA, \lambda\in[0,1]\} \\
    &= \{(1-\lambda)Q(\os, \oa_1)+\lambda Q(\os, \oa_2), \oa_1,\oa_2\in\ocA, \lambda\in[0,1]\} \\
    &= \cC^2(Q(\os,\ocA))\}
\end{align*}

Hence, the problem \eqref{eq:pi_greedy_reward}, \eqref{eq:pi_greedy_constraint} has become:
\begin{equation*}
    \tilde{\Pi}^Q_r = \argmax_{(q_r, q_c)\in\cC^2(Q(\os, \ocA))} q_r \quad\text{ s.t. }\quad q_c \leq \beta 
\end{equation*}
and the solution of $\pi_\text{greedy}$ is $q^*=\argmin_{q\in\tilde{\Pi}^Q_r} q_c$. 

The original problem in the space of actions $\ocA$ is now expressed in the space of values $Q(\os, \ocA)$ (which is why we use $=$ instead of $\in$ before $\argmin$ here).

We further restrict the search space of $q^*$ following two observations:
\begin{enumerate}
    \item $q^*$ belongs to the \emph{undominated} points $\cC^2(Q^-)$:
    \begin{align}
        \label{eq:q_minus_undominated}
        Q^+ &= \{(q_c, q_r): q_c > q_c^{\pm} = \min_{q^+} q_c^+\text{ s.t. }q^+\in\argmax_{q\in Q(\os,\ocA)} q_r\}\\
        Q^- &= Q(\os,\ocA) \setminus Q^+
    \end{align}
    Denote $q^*$ = $(1-\lambda) q^1 + \lambda q^2$, with $q^1, q^2\in Q(\os,\ocA)$. There are three possible cases:
    \begin{enumerate}
        \item $q^1, q^2 \not\in Q^-$. Then $q_c^* = (1-\lambda) q^1_c + \lambda q^2_c > q_c^{\pm}$. But then $q_c^{\pm} < q_c^* \leq \beta$ so $q^{\pm}\in\tilde{\Pi}^Q_r$ with a strictly lower $q_c$ than $q^*$, which contradicts the $\argmin$.
        \item $q^1\in Q^-, q^2 \not\in Q^-$. But then consider the mixture $q^\top = (1-\lambda) q^1 + \lambda q^\pm$. Since $q_r^{\pm} \geq q_r^{2}$ and $q_r^{\pm} < q_r^{2}$, we also have $q^\top_r \geq q_r^*$ and $q^\top_c < q_c^*$, which also contradicts the $\argmin$.
        \item $q^1,q^2\in Q^-$ is the only remaining possibility.
    \end{enumerate}
    \item $q^*$ belongs to the \emph{top frontier} $\cF$:
    \begin{equation*}
        \cF_Q = \{q\in \cC^2(Q^-): \not\exists q'\in \cC^2(Q^-): q_c=q_c'\text{ and }q_r<q_r'\}
    \end{equation*}
    Trivially, otherwise q' would be a better candidate than $q^*$.
\end{enumerate}

Let us characterise this frontier $\cF$. It is both:
\begin{enumerate}
    \item the \emph{graph of a non-decreasing function}: $\forall q^1, q^2\in\cF$ such that $q_c^1\leq q_c^2$ then $q_r^1\leq q_r^2$.\\
    By contradiction, if we had $q_r^1 > q_r^2$, we could define $q^\top = (1-\lambda)q^1 + \lambda q^\pm$ where $q^\pm$ is the dominant point as defined in \eqref{eq:q_minus_undominated}. By choosing $\lambda=(q^2_c-q^1_c)/(q^\pm_c-q^1_c)$ such that $q^\top_c = q_c^2$, then since $q_r^\pm \geq q_r^1 > q_r 2$ we also have $q^\top_r > q_r^2$ which contradicts $q^2\in\cF$.
    \item the \emph{graph of a concave function}: $\forall q^1, q^2, q^3\in\cF$ such that $q_c^1\leq q_c^2 \leq q_c^3$ with $\lambda$ such that $q^2_c = (1-\lambda)q^1_c + \lambda q^3_c$, then $q_r^2 \geq (1-\lambda)q_r^1 + \lambda q_r^3$.\\
    Trivially, otherwise the point $q^\top = (1-\lambda)q^1 + \lambda q^3$ would verify $q^\top_c=q^2_c$ and $q^\top_r > q^2_r$, which would contradict $q^2 \in\cF$.
\end{enumerate}

We denote $\cF_Q = \cF \cap Q$. Clearly, $q^*\in\cC^2(\cF_Q)$: let $q^1, q^2\in Q^-$ such that $q^* = (1-\lambda)q^1  + \lambda q^2$. First, $q^1, q^2\in Q^-\subset\cC^2(Q^-)$. Then, by contradiction, if there existed $q^{1'}$ or $q^{2'}$ with equal $q_c$ and strictly higher $q_r$, again we could build an admissible mixture $q^{\top}=(1-\lambda)q^{1'}  + \lambda q^{2'}$ strictly better than $q^*$.

$q^*$ can be written as $q^* = (1-\lambda)q^1  + \lambda q^2$ with $q^1, q^2\in\cF_Q$ and, without loss of generality, $q^1_c \leq q^2_c$.

\textbf{Regular case:} there exists $q^0\in\cF_Q$ such that $q^0_c \geq \beta$.

Then $q^1$ and $q^2$ must flank the budget: $q_c^1 \leq \beta \leq q_c^2$. Indeed, by contradiction, if $q_c^2 \geq q_c^1 > \beta$ then $q_c^* > \beta$ which contradicts $\Pi_r^Q$. Conversely, if $q_c^1 \leq q_c^2 < \beta$ then $q^* < \beta \leq q^0_c$, which would make $q^*$ a worse candidate than $q^\top=(1-\lambda)q^* + \lambda q^0$ when $\lambda$ is chosen such that $q_c^\top=\beta$, and contradict $\Pi_r^Q$ again.

Because $\cF$ is the graph of a non-decreasing function, $\lambda$ should be as high as possible, as long as the budget $q^*\leq\beta$ is respected. We reach the highest $q_r^*$ when $q^*_c=\beta$, that is: $\lambda=(\beta-q_c^1)/(q_c^2-q_c^1)$.

It remains to show that $q^1$ and $q^2$ are two successive points in $\cF_Q$: $\not\exists q\in\cF_Q\setminus\{q^1, q^2\}: q^1_c \leq q_c \leq q^2_c$. Otherwise, as $\cF$ is the graph of a concave function, we would have $q_r \geq (1-\mu)q_r^1 + \mu q_r^2$. $q_r$ cannot be strictly greater than $(1-\mu)q_r^1 + \mu q_r^2$ which would contradict $q^*$, but it can still be equal, which means the tree points $q, q^1, q^2$ are aligned. In fact, every points aligned with $q^1$ and $q^2$ can also be used to construct mixtures resulting in $q^*$, but among these solutions we can still choose $q^1$ and $q^2$ as the two points in $\cF_Q$ closest to $q^*$.

\textbf{Edge case:} $\forall q\in\cF_Q, q_c < \beta$. Then  $q^* =  \argmax_{q\in\cF} q_r = q^\pm =  \argmax_{q\in Q^-} q_r$

\end{proof}

\section{Risk-Sensitive Exploration}
\label{sec:risk-sensitive-supp}
We recall the Risk-Sensitive Exploration algorithm in \Cref{algo:risk-sensitive-exploration}

\begin{algorithm}[ht!]
\DontPrintSemicolon
\KwData{An environment, a BFTQ solver, $W$ CPU workers}
\KwResult{A batch of transitions $\cD$}
$\cD\leftarrow\{\}$\;
\For{each intermediate batch} {
split episodes between $W$ workers\;
\For(\tcp*[f]{run this loop on each worker in parallel}){each episode in batch}{
sample initial budget $\beta\sim\mathcal{U}(\mathcal{B})$.\;
\While{episode not done}{
update $\epsilon$ from schedule.\;
sample $z\sim\mathcal{U}([0, 1])$.\;
\lIf{$z < \epsilon$}{sample $(a, \beta_a)\sim\mathcal{U}(\Delta_{\cA\cB})$.\tcp*[f]{Explore}}
\lElse{sample $(a, \beta_a)\sim\pi_\text{greedy}(a, \beta_a|s, \beta; Q^*)$.\tcp*[f]{Exploit}}
append transition $(s, \beta, a, \beta_a, R, C, s')$ to batch $\mathcal{D}$.\;
step episode budget $\beta \leftarrow \beta_a$
}
}
$\pi_\text{greedy}(\cdot\sim; ~Q^*) \leftarrow\texttt{BFTQ}(\cD)$.
}
\Return{the batch of transitions $\cD$}
\caption{Risk-sensitive exploration}
\label{algo:risk-sensitive-exploration}
\end{algorithm}

\section{Scalable Implementation of BFTQ}
\label{sec:bftq-full}

We recall the scalable version of BFTQ in \Cref{algo:bftq_full} and the architecture of the neural network \Cref{fig:architecture}.

\begin{figure}[tp]
    \centering
    \begin{tikzpicture}[shorten >=1pt,->,draw=black!50, inner sep=1pt, node distance=\layersep]
        \tikzstyle{every pin edge}=[<-,shorten <=1pt]
        \tikzstyle{neuron}=[circle,fill=black!25,minimum size=10pt,inner sep=0pt]
        \tikzstyle{input neuron}=[neuron];
        \tikzstyle{input beta}=[neuron];
        \tikzstyle{qc}=[neuron];
        \tikzstyle{qr}=[neuron];
        \tikzstyle{hidden neuron}=[neuron];
        \tikzstyle{autoencoder neuron}=[neuron];
        \tikzstyle{annot} = [text width=4em, text centered]
        \tikzstyle{annot2} = [text width=10em, text centered]
        \def\layersep{2cm}
        \foreach \name / \y in {0,...,1}
            \pgfmathtruncatemacro{\y}{0 + \y}
            \node[input neuron, pin=left:$s^\y$] (I-\name) at (0,-\y) {};

        \foreach \name / \y in {0,...,4}
        \pgfmathtruncatemacro{\y}{0 + \y}
            \path node[hidden neuron] (H1-\name) at (1*\layersep,-\y cm) {};

        \foreach \source in {0,...,1}
            \foreach \dest in {0,...,4}
                \path (I-\source) edge (H1-\dest);

        \node[input beta, pin=left:$\beta_a$] (BETA) at (0,-3.5) {};

        \foreach \name / \y in {0,...,1}
            \pgfmathtruncatemacro{\ybis}{3+ \y}
            \path node[autoencoder neuron] (AE-\name) at (\layersep/3.5,-\ybis cm) {};

        \foreach \name / \y in {0,...,3}
        \pgfmathtruncatemacro{\y}{0 + \y}
            \path[yshift=-0.5cm]
            node[hidden neuron] (H2-\name) at (1.6*\layersep,-\y cm) {};

        \foreach \name / \y in {0,...,1}
        \pgfmathtruncatemacro{\y}{0 + \y}
            \path[yshift=-0.5cm] node[qr,pin=right:$Q_r(a^\y)$] (Qr-\name) at (2.2*\layersep,-\y cm) {};

        \foreach \name / \y in {0,...,1}
        \pgfmathtruncatemacro{\yy}{2 + \y}
            \path[yshift=-0.5cm] node[qc,pin=right:$Q_c(a^\y)$] (Qc-\name) at (2.2*\layersep,-\yy cm) {};

        \foreach \source in {0,...,1}
            \foreach \dest in {0,...,4}
                \path (AE-\source) edge (H1-\dest);

        \foreach \dest in {0,...,1}
            \path (BETA) edge (AE-\dest);

         \foreach \source in {0,...,4}
            \foreach \dest in {0,...,3}
                \path (H1-\source) edge (H2-\dest);

        \foreach \source in {0,...,3}
            \foreach \dest in {0,...,1}
                \path (H2-\source) edge (Qr-\dest);

         \foreach \source in {0,...,3}
            \foreach \dest in {0,...,1}
                \path (H2-\source) edge (Qc-\dest);
       \node[annot] (input) at (0,1) {$(s,\beta_a)$};
       \node[annot2] (input) at (\layersep/3.5,-4.6) {Encoder};
       \node[annot](h1) at (\layersep,1) {Hidden Layer 1};
       \node[annot](h2) at(1.6* \layersep,1) {Hidden Layer 2};
       \node[annot](output) at(2.2* \layersep,1) {$Q$};
    \end{tikzpicture}
    \caption{Neural Network for $Q$-functions approximation when $\cS=\Real^2$ and $|\cA| = 2$.}
    \label{fig:architecture}
\end{figure}

\begin{algorithm}[tp]
\DontPrintSemicolon
\KwData{$\mathcal{D}$, $\tilde{\mathcal{B}}$ a finite subset of $\mathcal{B}$, $\gamma$, a model $Q\in (\Real^2)^{S \ocA}$, a regression algorithm \texttt{fit}, a set of CPU workers $W$}
\KwResult{$Q^*$}
$Q \leftarrow 0$\;
$X \leftarrow \{s_i,a_i,\beta_{a_i}\}_{i\in[0, |\cD|]}$\;
$S' \leftarrow \{s_i'\}_{i\in[0, |\cD|]}$\;
\Repeat{convergence}{
   Evaluate $Q(S', \cA, \tilde{\cB})$ in a single forward pass\;
   Split $\mathcal{D}$ among workers: $\mathcal{D} = \cup_{w\in W} \mathcal{D}_w$\;
   \For(\tcp*[f]{Run in parallel}){$w\in W$}{
       \For{$(\boldsymbol{\cdot},\boldsymbol{\cdot},\beta_{a_i},{R_r}_i,{R_c}_i,s'_i) \in \mathcal{D}$} {
           $\cP \leftarrow \{(Q_c(s_i',\cA,\tilde{\cB}), Q_r(s_i',\cA,\tilde{\cB}))\}$\;
           $\cP.\texttt{prune}()$ \tcp*[f]{Remove all dominated points}\;
           $\cH \leftarrow \texttt{convex\_hull}(\cP).\text{vertices}()$\tcp*[f]{in cw order}\;
           $k \leftarrow \min\{k: \beta_i \geq q_c$ with $\left(q_c,q_r\right) = \cH[k]\}$\;
           $q_c^2,q_r^2,q_c^1,q_r^1 \leftarrow \cH[k],\cH[k-1]$\;
           $p \leftarrow (\beta_{a_i} - q_a^1) / (q_c^2 - q_c^1)$\;
           $Y_c^{w,i} \leftarrow {R_c}_i + \gamma ((1-p) q_c^1+ p q_c^2)$\;
           $Y_r^{w,i} \leftarrow {R_r}_i + \gamma ((1-p) q_r^1+ p q_r^2)$\;
       }
   }
   Join the results: $Y \leftarrow \cup_{w\in W} (Y_c^w, Y_r^w)$\;
   $Q \leftarrow \texttt{fit}(X, Y)$\;
}
\caption{Scalable BFTQ}
\label{algo:bftq_full}
\end{algorithm}

\section{The Lagrangian Relaxation Baseline}
\label{sec:lagragian}
\begin{figure}[tp]
    \centering
    \includegraphics[width=0.5\textwidth]{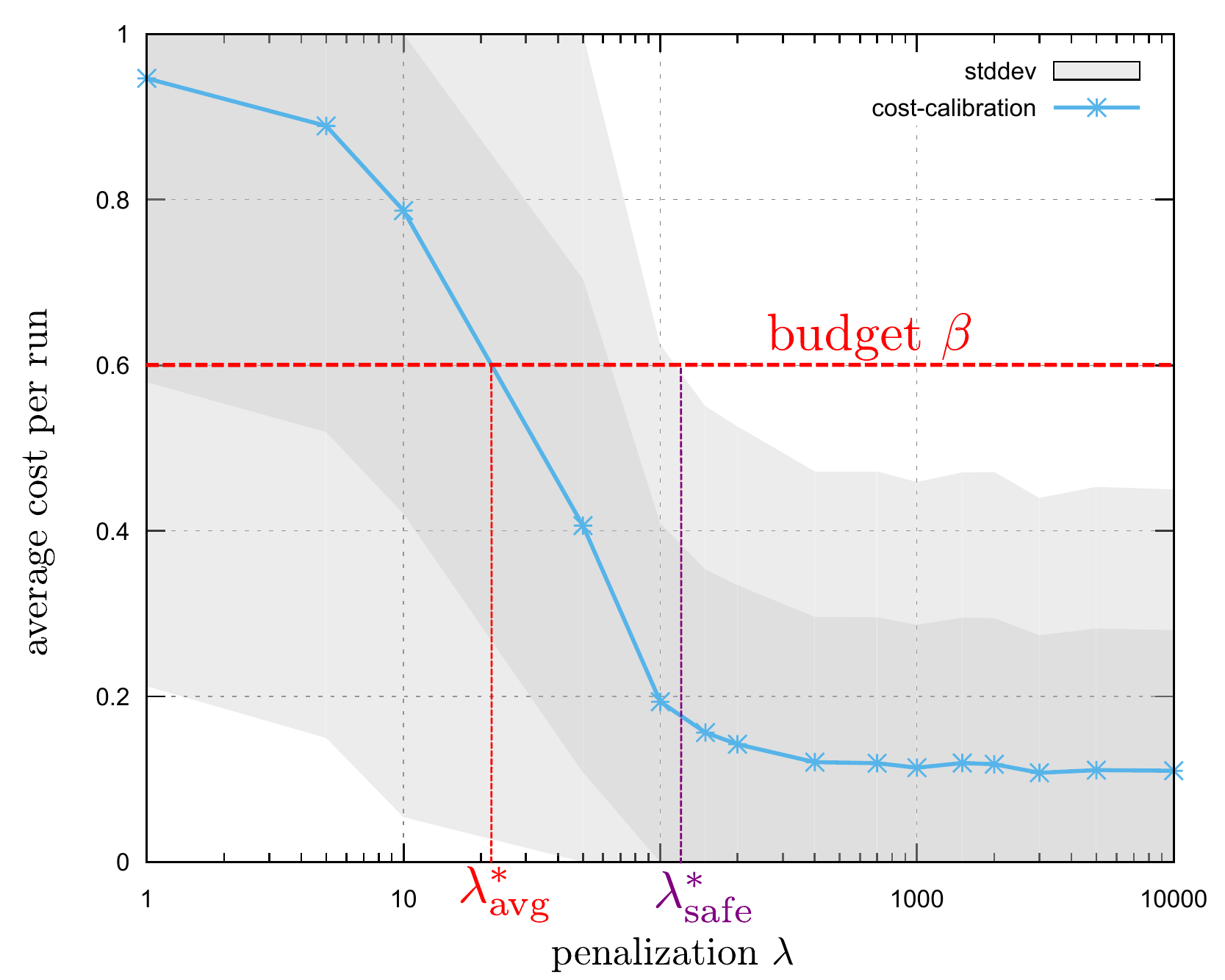}
    \caption{Calibration of a penalty multiplier according to the budget $\beta$. The optimal multiplier $\lambda^*_{\text{avg}}$ is the smallest one to satisfy the budget constraint on average. Safer policies can also be selected according to the largest deviation from this mean cost.}
    \label{fig:Lagrangian}
\end{figure}

As explained on \Cref{fig:Lagrangian}, the optimal deterministic policy can be obtained by a line-search on the Lagrange multiplier values $\lambda$.

Then, according to \citet[Theorem 4.4]{BEUTLER1985236}, the optimal policy is a randomised mixture of two deterministic policies: the safest deterministic policy that violates the constraint $\pi_{\lambda-}$ and the riskier of the feasible ones $\pi_{\lambda+}$.

Fitted-Q (FTQ)~\citep{Ernst2005,Riedmiller2005} can be easily adapted for continuous states CMDP and BMDP through this methodology, but given the high variance it requires a lot of simulations to get a proper estimate of the calibration curve. Our purpose is to avoid this calibration phase.

\section{Experiments}
\label{sec:exp-supp}

\subsection{Examples of different exploration strategies}
\label{subsec:exploration-examples}
We compare two approaches for constructing a batch of samples. The animations from the html page \url{exploration.html} display the trajectories collected in each intermediate sub-batch. The first row corresponds to a classical  risk neutral epsilon-greedy exploration policy while the second row showcases a risk-sensitive exploration strategy introduced in the paper. Each animation corresponds to a different seed.

\subsection{Examples of BFTQ policies executions}
\label{sec:bftq-executions}

We display the evolution in the budgeted policy behaviour with respect to the budget on different environments. The policies have been learnt with a risk-sensitive exploration.

\paragraph{Highway-Env}

On the \texttt{highway-env} , the budgeted agents display a wide variety of behaviours. Animations are displayed on the html page \url{highway-env.html}. When $\beta = 1$, the ego-vehicle drives in a very aggressive style: it immediately switches to the opposite lane and drives as fast as possible to pass slower vehicles, swiftly changing lanes to avoid incoming traffic. On the contrary when $\beta = 0$, the ego-vehicle is conservative: it stays on its lane and drives at a low velocity. With intermediate budgets such as $\beta = 0.2$, the agent sometimes decides to overtake its front vehicle but promptly steers back to its original lane afterwards.

\paragraph{Slot-filling}

\paragraph{Remark on the \texttt{slot-filling} environment} When receiving an utterance, the system can either understand it $(\mu=\mu_u)$ or misunderstand it $(\mu=\mu_m)$ with a fixed probability called the sentence error rate $ser$. Then, the speech recognition score is simulated \citep{Khouzaimi2015}: $srs = (1+\exp(-x))^{-1}$ with $x\sim N(\mu, \sigma)$. It's the confidence score of the natural language understanding module about the last utterance. Note that here are no recognition errors ($ser=0$ and $srs=1$) when the user provides information using the numeric pad.

In \Cref{table:dialogues}, we display two dialogues done with the same BFTQ policy on \texttt{slot-filling}. The policy is given two budgets to respect in expectation, $\beta=0$ and $\beta=0.5$. For $\beta=0$, one can see that the system never uses the \texttt{ask\_num\_pad} action. Instead, it uses \texttt{ask\_oral} , an action subject to recognition errors. The system keeps asking for the same slot 2, because it has the lowest speech recognition score. It eventually summarises the form to the user, but then reaches the maximum dialogue length and thus faces a dialogue failure. For $\beta=0.5$, the system first asks in a safe way, with \texttt{ask\_oral}. It may want to \texttt{ask\_num\_pad} if one of the speech recognition score is low. Then, the system proceeds to a confirmation of the slot values. If it is incorrect, the system continues the dialogue using unsafe the \texttt{ask\_num\_pad} action to be certain of the slot values.

\begin{table}[tp]
\centering
\resizebox{\textwidth}{!}{
\begin{tabular}[]{lll}
\toprule

turn&$\beta=0$&$\beta=0.5$\tabularnewline
\midrule
turn 0& \makecell[l]{valid slots : [0, 0, 0]\\ srs : [ None None None ]\\ system says ASK\_ORAL(1) \\ user says INFORM} &\makecell[l]{ valid slots : [0, 0, 0]\\ srs : [ None None None ]\\ system says ASK\_ORAL(2)\\ user says INFORM}\tabularnewline\midrule
turn 1&\makecell[l]{valid slots : [0, 0, 0]\\ srs : [ None 0.48 None ]\\ system says ASK\_ORAL(2) \\ user says INFORM} & \makecell[l]{valid slots : [0, 0, 1]\\ srs : [ None None 0.56 ]\\ system says ASK\_ORAL(0)\\ user says INFORM}\tabularnewline\midrule
turn 2&\makecell[l]{valid slots : [0, 0, 0]\\ srs : [ None 0.48 0.22 ]\\ system says ASK\_ORAL(0) \\ user says INFORM} & \makecell[l]{valid slots : [0, 0, 1] \\ srs : [ 0.30 None 0.56 ]\\ system says ASK\_ORAL(1)\\ user says INFORM}\tabularnewline\midrule
turn 3&\makecell[l]{valid slots : [0, 0, 0]\\ srs : [ 0.62 0.48 0.22 ]\\ system says ASK\_ORAL(2) \\ user says INFORM} & \makecell[l]{valid slots : [0, 0, 1]\\ srs : [ 0.30 0.54 0.56 ]\\ system says ASK\_ORAL(0)\\ user says INFORM}\tabularnewline\midrule
turn 4&\makecell[l]{valid slots : [0, 0, 0] \\ srs : [ 0.62 0.48 0.66 ]\\ system says ASK\_ORAL(1) \\ user says INFORM} & \makecell[l]{valid slots : [0, 0, 1]\\ srs : [ 0.68 0.54 0.56 ]\\ system says ASK\_NUM\_PAD(1)\\ user says INFORM}\tabularnewline\midrule
turn 5&\makecell[l]{valid slots : [0, 1, 0]\\ srs : [ 0.62 0.56 0.66 ]\\ system says ASK\_ORAL(2) \\ user says INFORM} & \makecell[l]{valid slots : [0, 1, 1]\\ srs : [ 0.68 1.00 0.56 ]\\ system says SUMMARIZE\_AND\_INFORM\\ user says DENY\_SUMMARIZE}\tabularnewline\midrule
turn 6&\makecell[l]{valid slots : [0, 1, 0]\\ srs : [ 0.62 0.56 0.14 ]\\ system says ASK\_ORAL(2) \\ user says INFORM} & \makecell[l]{valid slots : [0, 1, 1]\\ srs : [ 0.68 1.00 0.56 ]\\ system says ASK\_NUM\_PAD(2)\\ user says INFORM}\tabularnewline\midrule
turn 7&\makecell[l]{valid slots : [0, 1, 1]\\ srs : [ 0.62 0.56 0.30 ]\\ system says ASK\_ORAL(2) \\ user says INFORM }& \makecell[l]{valid slots : [0, 1, 1]\\ srs : [ 0.68 1.00 1.00 ]\\ system says SUMMARIZE\_AND\_INFORM\\ user says DENY\_SUMMARIZE}\tabularnewline\midrule
turn 8&\makecell[l]{valid slots : [0, 1, 1]\\ srs : [ 0.62 0.56 0.49 ]\\ system says ASK\_ORAL(2) \\ user says INFORM} & \makecell[l]{valid slots : [0, 1, 1]\\ srs : [ 0.68 1.00 1.00 ]\\ system says ASK\_NUM\_PAD(0)\\ user hangs up !}\tabularnewline\midrule
turn 9&\makecell[l]{valid slots : [0, 1, 1]\\ srs : [ 0.62 0.56 0.65 ]\\ system says SUMMARIZE\_AND\_INFORM \\ max size reached !}& \makecell[l]{ }\tabularnewline\bottomrule
\end{tabular}
}
    \caption{Two dialogues generated by a safe policy ($\beta=0$) on the left and a risky one ($\beta=0.5$) on the right.}
    \label{table:dialogues}
\end{table}

\paragraph{Corridors}

Animations are displayed on the html page \url{corridors.html} for the \texttt{corridors} environment. When the budget is low, the agent takes the safest path on the left. When the budget increases, it gradually switches to the other lane, earning higher rewards but also costs. This gradual process could not be achieved with a deterministic policy as it would chose either one path or the other. Each animation corresponds to a different seed.

\subsection{Reproducibility}
\label{subsec:reproducibility-supp}

The following section displays environments and algorithms parameters and instructions to reproduce the exact same results displayed in \Cref{sec:experiements}.

\subsubsection{Environments Parameters}
\label{sec:env-parameters}

All environments parameters are displayed in \Cref{tab:param-corridors}, \Cref{tab:param-slot-filling} and \Cref{tab:param-highway-env}.

\paragraph{State-Space}

The states $s$ (from $\os=(s,\beta)$) of the agent are described in the following:

\begin{itemize}
    \item \texttt{Corridors}: $s = (x,y)$ where $x$ and $y$ are the 2D coordinates of the agent.
    \item \texttt{Slot-Filling}: $s = (\text{srs},\text{min},a_u,a_s,t)$ where $\text{srs}$ is a vector of the speech recognition score for each slot, $\text{min}$ is a one hot vector describing the minimum of the $\text{srs}$ vector, $a_u$ is a one hot vector of the last user dialogue act and $a_s$ is the one hot vector of the last system dialogue act. Finally $t\in[0,1]$ is the fraction of the current turn with the maximum number of turns authorised.
    \item \texttt{Highway-Env}: the positions $(x, y)$ and velocities $(\dot{x}, \dot{y})$ of every vehicle on the road.
\end{itemize}

\begin{table}[ht!]
    \centering
    \begin{tabularx}{1.0\textwidth}{lll}
        \toprule
        Parameter & Description & Value\tabularnewline
        \midrule
        - & Size of the environment & 7 x 6\tabularnewline
        - & \makecell[l]{Standard deviation of the Gaussian \\noise applied to actions} & (0.25,0.25)\tabularnewline
        H & Episode duration & 9\tabularnewline
        \bottomrule
    \end{tabularx}
    \caption{Parameters of \texttt{Corridors}}
    \label{tab:param-corridors}
\end{table}

\begin{table}[ht!]
    \centering
    \begin{tabularx}{1.0\textwidth}{lll}
        \toprule
        Parameter & Description & Value\tabularnewline
        \midrule
        ser & Sentence Error Rate & 0.6\tabularnewline
        $\mu_m$& Gaussian mean for misunderstanding & -0.25\tabularnewline
        $\mu_u$& Gaussian mean for understanding & 0.25\tabularnewline
        $\sigma$& Gaussian standard deviation & 0.6\tabularnewline
        $p$& Probability of hang-up & 0.25\tabularnewline
        H & Episode duration & 10\tabularnewline
        - & Number of slots & 3\tabularnewline
        \bottomrule
    \end{tabularx}
    \caption{Parameters of \texttt{Slot-Filling}}
    \label{tab:param-slot-filling}
\end{table}

\begin{table}[ht!]
    \centering
    \begin{tabularx}{1.0\textwidth}{lll}
        \toprule
        Parameter & Description & Value\tabularnewline
        \midrule
        $N_v$& Number of vehicles & 2 - 6\tabularnewline
        $\sigma_p$& Standard deviation of vehicles initial positions & 100 m\tabularnewline
        $\sigma_v$& Standard deviation of vehicles initial velocities & 3 m/s\tabularnewline
        H & Episode duration & 15 s\tabularnewline
        \bottomrule
    \end{tabularx}

    \caption{Parameters of \texttt{highway-env}}
    \label{tab:param-highway-env}
\end{table}

\subsubsection{Algorithm parameters}
\label{sec:algorithms-parameters}

All algorithm parameters are displayed in \Cref{tab:param-algo-corridors},\Cref{tab:param-algo-slot-filling} and \Cref{tab:param-algo-highway-env}.

\paragraph{A note on the parameters search}

We performed a shallow grid-search for the classic Neural-Network parameters. Most of the parameters don't have a strong influence on the results, however in the \texttt{slot-filling} environment, the choice of the regulation weight is decisive.

\begin{table}[tp]
    \centering
    \begin{tabularx}{1.0\textwidth}{lll}
        \toprule
        Parameters & BFTQ(risk-sensitive) & BFTQ(risk-neutral)\tabularnewline
        \midrule
        architecture & 256x128x64 & 256x128x64\tabularnewline
        regularisation & 0.001 & 0.001\tabularnewline
        activation & relu & relu\tabularnewline
        size beta encoder & 3 & 3\tabularnewline
        initialisation & xavier & xavier\tabularnewline
        loss function & L2 & L2\tabularnewline
        optimizer & adam & adam\tabularnewline
        learning rate & 0.001 & 0.001\tabularnewline
        epoch (NN) & 1000 & 5000\tabularnewline
        normalize reward & true & true\tabularnewline
        epoch (FTQ) & 12 & 12\tabularnewline
        $\tilde{\cB}$ & 0:0.01:1 & -\tabularnewline
        $\gamma$ & 1 & 1\tabularnewline
        $N=|\cD|$ & 5000 & 5000\tabularnewline
        $N_\text{minibatch}$ & 10 & 10\tabularnewline
        $N_\text{seeds}$ & 4 & 4\tabularnewline
        $N_\text{test}$ & 1000 & 1000\tabularnewline
        decay epsilon scheduling & 0.001 & 0.001\tabularnewline
        \bottomrule

    \end{tabularx}
    \caption{Algorithms parameters for \texttt{Corridors}}
    \label{tab:param-algo-corridors}
\end{table}

\begin{table}[tp]
    \centering
    \begin{tabularx}{1.0\textwidth}{lll}
        \toprule
        Parameters & BFTQ & FTQ\tabularnewline
        \midrule
        architecture & 256x128x64 & 128x64x32\tabularnewline
        regularisation & 0.0005 & 0.0005\tabularnewline
        activation & relu & relu\tabularnewline
        size beta encoder & 50 & -\tabularnewline
        initialisation & xavier & xavier\tabularnewline
        loss function & L2 & L2\tabularnewline
        optimizer & adam & adam\tabularnewline
        learning rate & 0.001 & 0.001\tabularnewline
        epoch (NN) & 5000 & 5000\tabularnewline
        normalize reward & true & true\tabularnewline
        epoch (FTQ) & 11 & 11\tabularnewline
        $\tilde{\cB}$ & 0:0.01:1 & -\tabularnewline
        $\gamma$ & 1 & 1\tabularnewline
        $N=|\cD|$ & 5000 & 5000\tabularnewline
        $N_\text{minibatch}$ & 10 & 10\tabularnewline
        $N_\text{seeds}$ & 6 & 6\tabularnewline
        $N_\text{test}$ & 1000 & 1000\tabularnewline
        decay epsilon scheduling & 0.001 & 0.001\tabularnewline
        \bottomrule
    \end{tabularx}
    \caption{Algorithms parameters for \texttt{Slot-Filling}}
    \label{tab:param-algo-slot-filling}
\end{table}
\begin{table}[tp]
    \centering
    \begin{tabularx}{1.0\textwidth}{lll}
        \toprule
        Parameters & BFTQ & FTQ\tabularnewline
        \midrule
        architecture & 256x128x64 & 128x64x32\tabularnewline
        regularisation & 0.0005 & 0\tabularnewline
        activation & relu & relu\tabularnewline
        size beta encoder & 50 & -\tabularnewline
        initialisation & xavier & xavier\tabularnewline
        loss function & L2 & L2\tabularnewline
        optimizer & adam & adam\tabularnewline
        learning rate & 0.001 & 0.01\tabularnewline
        epoch (NN) & 5000 & 400\tabularnewline
        normalize reward & true & true\tabularnewline
        epoch (FTQ) & 15 & 15\tabularnewline
        $\tilde{\cB}$ & 0:0.01:1 & -\tabularnewline
        $\gamma$ & 0.9 & 0.9\tabularnewline
        $N=|\cD|$ & 10000 & 10000\tabularnewline
        $N_\text{minibatch}$ & 10 & 10\tabularnewline
        $N_\text{seeds}$ & 10 & 10\tabularnewline
        $N_\text{test}$ & 150 & 150\tabularnewline
        decay epsilon scheduling & 0.0003 & 0.0003\tabularnewline
        \bottomrule
    \end{tabularx}
    \caption{Algorithms parameters for \texttt{Highway-Env}}
    \label{tab:param-algo-highway-env}
\end{table}

\subsubsection{Instructions for reproducibility}
\label{subsubsec:instruction-reproducibility}
To reproduce the result displayed in \Cref{sec:experiements}, first install the following conventional libraries for python3: pycairo, numpy, scipy and pytorch. Then, execute the commands in \Cref{fig:instructions} on a Linux Operating System. The Graphic Processing Unit used for experiments is an NVIDIA \texttt{GeForce GTX 1080 Ti} and the Computational Processing Unit is an Intel \texttt{Xeon E7}.

\begin{figure}
    \centering

\begin{lstlisting}[language=bash,caption={bash version}]
# Install highway-env
pip3 install --user git+https://github.com/eleurent/rl-agents
# Change python path to the path of this repository
export PYTHONPATH="code/scaling-up-brl"
# Navigate to budgeted-rl folder
cd code/scaling-up-brl/budgeted-rl/
# Run main script using any config file
# Choose the range of seeds you want to test on
python3 main/egreedy/main-egreedy.py config/slot-filling.json 0 6
python3 main/egreedy/main-egreedy.py config/corridors.json 0 4
python3 main/egreedy/main-egreedy.py config/highway-easy.json 0 10
\end{lstlisting}

\caption{Instructions to reproduce experiments}
    \label{fig:instructions}
\end{figure}

\section{The machine learning reproducibility checklist}
\label{sec:ml-checklist}
For all models and algorithms presented, indicate if you include: 

\begin{itemize}
    \item  A clear description of the mathematical setting, algorithm, and/or model:
    
    \begin{itemize}\item \textbf{yes}, see \Cref{sec:intro}, \Cref{sec:bdp}, \Cref{sec:brl},
    \Cref{sec:risk-sensitive-supp} and \Cref{sec:bftq-full}.
    \end{itemize}
    \item An analysis of the complexity (time, space, sample size) of any algorithm:
    \begin{itemize}\item \textbf{yes}, see \Cref{subsec:parallel-computing}.
    \end{itemize}
    \item A link to a downloadable source code, with specification of all dependencies, including external libraries:
    \begin{itemize}\item \textbf{yes}, see \Cref{subsubsec:instruction-reproducibility} and the folder \texttt{code} in the supplementary material zip file.
    \end{itemize}
\end{itemize}

For any theoretical claim, indicate if you include:

\begin{itemize}
    \item  A statement of the result:
     \begin{itemize}
        \item  \textbf{yes}, see \Cref{sec:bdp} and \Cref{sec:brl}.\end{itemize}
    \item A clear explanation of any assumptions:
    \begin{itemize}
        \item  we make one assumption in \Cref{sec:bdp}. We assume the program is feasible for any state. If not, no algorithm would be able to solve it anyway.\end{itemize}
    \item A complete proof of the claim:
    \begin{itemize}
        \item  \textbf{yes}, see \Cref{sec:proofs}. We formulate a conjecture in \Cref{rmk:contractivity-smooth} but we provide a sketch of the proof in \Cref{proof:contraction-with-smooth}.\end{itemize}
\end{itemize}

For all figures and tables that present empirical results, indicate if you include:

\begin{itemize}
    \item A complete description of the data collection process, including sample size: 
    
        \begin{itemize}
            \item\textbf{yes}, see \Cref{sec:experiements} and \Cref{sec:algorithms-parameters}.
        \end{itemize}
    \item A link to a downloadable version of the dataset or simulation environment:
    
        \begin{itemize}
            \item  \textbf{yes}, two environments   are shipped with the supplementary material (in the \texttt{code} folder) and the third one is fetch from a public repository, see \Cref{subsubsec:instruction-reproducibility} for details.
        \end{itemize}
    \item An explanation of any data that were excluded, description of any pre-processing step: 
        \begin{itemize}
            \item it's \textbf{not applicable} as data comes from simulated environments, so pre-processing steps are not needed.
        \end{itemize}

    \item An explanation of how samples were allocated for training / validation / testing: 
        \begin{itemize}
            \item it's \textbf{not applicable}. The complete dataset is used for training. There is no need for validation set. Testing is performed in the true environment as in classical online learning approaches.
        \end{itemize}

    \item The range of hyper-parameters considered, method to select the best hyper-parameter configuration, and specification of all hyper-parameters used to generate results: 
    
        \begin{itemize}
            \item \textbf{yes}, see \Cref{sec:algorithms-parameters}.
        \end{itemize}

    \item The exact number of evaluation runs: 
    
        \begin{itemize}
            \item \textbf{yes}, see $N_{seeds}$ in the tables from \Cref{sec:algorithms-parameters}.
        \end{itemize}

    \item A description of how experiments were run:              \begin{itemize}
            \item \textbf{yes}, see the two first paragraphs of \Cref{par:ex-explo}.
        \end{itemize}

    \item A clear definition of the specific measure or statistics used to report results:
        \begin{itemize}
            \item  \textbf{yes}, see \Cref{subsec:results}.
        \end{itemize}

    \item Clearly defined error bars:
    
        \begin{itemize}
            \item  \textbf{yes}, we plot 95\% confidence intervals in all figures, see \Cref{subsec:results}.
        \end{itemize}

    \item A description of results with central tendency (e.g. mean)  variation (e.g. stddev):
    
        \begin{itemize}
            \item  \textbf{yes}, we even observe less variability with our novel approach, see \Cref{subsec:results}.
        \end{itemize}
    \item A description of the computing infrastructure used:     \begin{itemize}
            \item The Graphic Processing Unit used for experiments is an NVIDIA \texttt{GeForce GTX 1080 Ti} and the Computational Processing Unit is an Intel \texttt{Xeon E7}.
        \end{itemize}
\end{itemize}

\end{document}